\documentclass[lettersize,journal]{IEEEtran}
\usepackage{amsmath,amsfonts}
\usepackage{amssymb}
\usepackage{algorithm}
\usepackage{array}
\usepackage[caption=false,font=normalsize,labelfont=sf,textfont=sf]{subfig}
\usepackage{textcomp}
\usepackage{stfloats}
\usepackage{url}
\usepackage{verbatim}
\usepackage{graphicx}
\usepackage{cite}
\usepackage{subfig}
\usepackage{gensymb}
\usepackage{algpseudocode}

\begin{document}

\title{An ocean front detection and tracking algorithm}

\author{
	\IEEEauthorblockN{Yishuo Wang$^{a,b}$, Feng Zhou$^{b*,a}$, Qicheng Meng$^{b}$, Muping Zhou$^{b}$, Zhijun Hu$^{c,b}$, Chengqing Zhang$^{d,b}$, Tianhao Zhao$^{d,b}$}
	
	\IEEEauthorblockA{$^a$ School of Oceanography, Shanghai Jiao Tong University, China}
	
	\IEEEauthorblockA{$^b$ State Key Laboratory of Satellite Ocean Environment Dynamics, Second Institute of Oceanography, MNR, China}
	
	\IEEEauthorblockA{$^c$ College of Marine Science and Technology, China University of Geosciences, China}
	
	\IEEEauthorblockA{$^d$ Ocean College, Zhejiang University, China}
	
	\IEEEauthorblockA{\{Yishuo Wang\}wys1998@sjtu.edu.cn, \{Feng Zhou\}zhoufeng@sio.org.cn, \{Zhijun Hu\}huzhijun@cug.edu.cn, \{Chengqing Zhang\}22234158@zju.edu.cn, \{Tianhao Zhao\}22334149@zju.edu.cn}
}

\maketitle

\begin{abstract}
Existing ocean front detection methods—including histogram-based variance analysis, Lyapunov exponent, gradient thresholding, and machine learning—suffer from critical limitations: discontinuous outputs, over-detection, reliance on single-threshold decisions, and lack of open-source implementations. To address these challenges, this paper proposes the Bayesian Front Detection and Tracking framework with Metric Space Analysis (BFDT-MSA). The framework introduces three innovations: (1) a Bayesian decision mechanism that integrates gradient priors and field operators to eliminate manual threshold sensitivity; (2) morphological refinement algorithms for merging fragmented fronts, deleting spurious rings, and thinning frontal zones to pixel-level accuracy; and (3) a novel metric space definition for temporal front tracking, enabling systematic analysis of front evolution. Validated on global sea surface temperature(SST) data (2022–2024), BFDT-MSA reduces over-detection by $73\%$ compared to histogram-based methods while achieving superior intensity ($0.16\degree$C/km), continuity, and spatiotemporal coherence. The algorithm's open-source release bridges a critical gap in reproducible oceanographic research.
\end{abstract}


\section{Introduction}
\IEEEPARstart{O}{cean} fronts, interfaces between distinct water masses, are pivotal in regulating marine ecosystems, heat transport, and extreme weather\cite{Woodson-2015, Champon-2020}. Despite advances in remote sensing, existing detection methods exhibit fundamental flaws\cite{Yang-2024}. Histogram-based approaches over-detect fragmented fronts due to rigid windowed F-tests, while Lagrangian methods (e.g., Lyapunov exponents) capture only eddy-associated fronts, ignoring wind- or topography-driven features\cite{Cayula-1992, Xing-2024, Nieto-2012, Xing-2023, Yang-2016, Aurell-1997, Prants-2014, Boffetta-2001, Prants-2011, Prants-2022, Tamim-2015}. Gradient thresholding, though widely adopted, produces discontinuous outputs and artificial rings, and machine learning merely replicates traditional results without innovation\cite{Ma-2024, Felt-2022, Felt-2023, Yang-2022, Thomas-2021, Canny-1986, Belkin-2009, Ren-2021, Artal-2019, Oram-2008}. Critically, no open-source framework exists for reproducible front tracking\cite{Yang-2024}.

This paper addresses these gaps through three key contributions:

1. Unified Bayesian Detection: By redefining fronts as probabilistic zones rather than binary edges, manual thresholds are replaced with a Bayesian fusion of gradient priors and field operators, enhancing robustness to noise and multi-scale variability.

2. Morphological Coherence Enforcement: A suite of algorithms—including ring deletion, branch trimming, and zone thinning—eliminates geometric artifacts while preserving topological continuity, a first in ocean front studies.

3. Metric Space Tracking: Front evolution analysis is formalized by defining temporal distances in a metric space, enabling rigorous tracking of front identity and dynamics over time.

The proposed BFDT-MSA framework is validated against four baseline methods: number, intensity, width and length, demonstrating superior performance in reducing over-detection ($71$ vs. $263$ fronts), sharpening intensity ($0.16\degree$C/km), and capturing realistic front dimensions . Its open-source release ensures transparency and adaptability for diverse applications, from climate modeling to fisheries management.

\subsection*{Key advantages over prior work}
Unlike threshold-dependent methods, BFDT-MSA leverages probabilistic reasoning to handle ambiguous gradients. Unlike Lagrangian approaches, it detects all front types—not just eddy-associated features. Unlike black-box machine learning, it provides interpretable, physics-guided outputs. These advancements establish a new paradigm for ocean front analysis.

\section{Related work}
In this section, a combination of gradient-based method, Bayesian decison and mathematical morphology is introduced, which can detect frontal zone and fronts. Further improvements with innovative algorithms are in the next section.

\subsection{Data}
The SST is obtained from the NOAA Coral Reef Watch daily global $5$km($0.05\degree$ exactly) satellite coral bleaching heat stress monitoring product(v3.1). It addresses errors identified in previous versions of the SST data, particularly for coral reef regions during the period from 2013-2016. What is more, this fine-grained resolution is essential for understanding ocean fronts and it can be downloaded from \url{https://coralreefwatch.noaa.gov/product/5km/}. 

Results of Lyapunov method for comparison are composed of multimission altimetry-derived gridded backward-in-time FSLE and Orientations of associated eigenvectors\cite{Hernandez-2011}. Spatial resolution is $4$km($0.04$ degree exactly) and can be downloaded from \url{https://www.aviso.altimetry.fr/en/data/products/value-added-products/fsle-finite-size-lyapunov-exponents.html}.

South China sea(SCS) is selected as the region studied in this paper. The spatial range is $100-125\degree$E and $0-25\degree$N, with the temporal range 2022-2024.

\subsection{Gradient calculation}
\begin{equation}\label{eq:kernel}
	\begin{gathered}
		F_{x}=\begin{pmatrix}-1&0&+1\\-2&0&+2\\-1&0&+1\end{pmatrix}\\
		F_{y}=\begin{pmatrix}+1&+2&+1\\0&0&0\\-1&-2&-1\end{pmatrix}
	\end{gathered}
\end{equation}

\begin{equation}
	\begin{gathered}
		T_x=F_x*SST\\
		T_y=F_y*SST
	\end{gathered}
\end{equation}

The $F_x$ and $F_y$ in \eqref{eq:kernel} are two convolutional kernels which perform weighted operations on the surrounding points of each point, respectively, to calculate the longitudinal gradient $T_y$ and latitudinal gradient $T_x$.

\begin{equation}\label{eq:mag_dir}
	\begin{gathered}
		T_{mag}=\sqrt{T_x^2+T_y^2}\\
		T_{dir}=arctan(\frac{T_y}{T_x})
	\end{gathered}
\end{equation}

Using $T_x$ and $T_y$, the magnitude and direction of gradient can be calculated by \eqref{eq:mag_dir}. The range of gradient direction is $[-\pi, \pi)$.

\subsection{Double thresholding}
\begin{figure}[!t]
	\centering
	\includegraphics[width=0.35\textwidth]{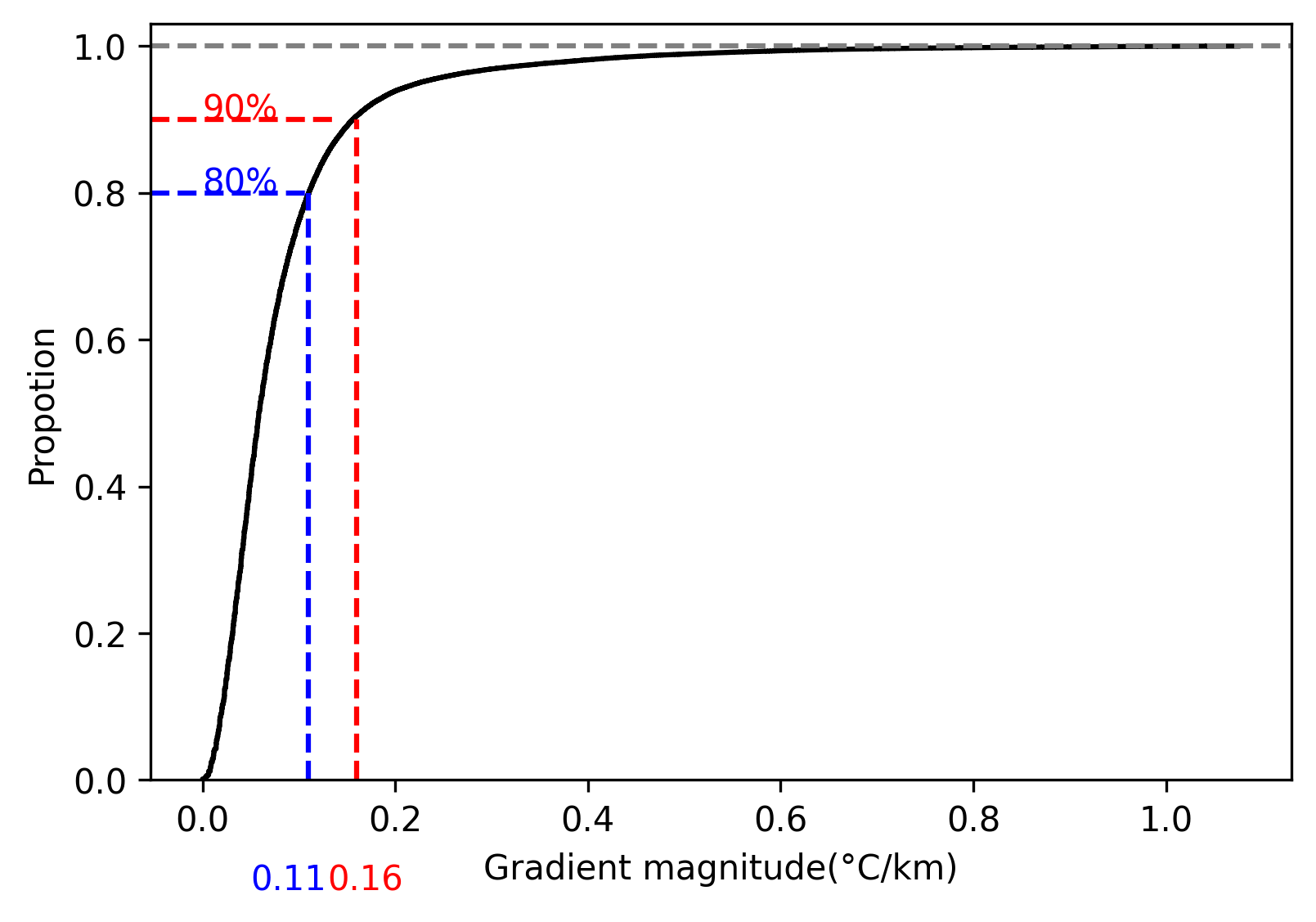}
	\caption{CDF of SST gradient magnitude in 2025.01.01} 
	\label{fig:cdf} 
\end{figure}

Sort according to $T_{mag}$ in descending order to generate a culmulative distribution function(CDF). Then, take the top 10\% $T_{mag}$ position in the gradient sequence as the first percentile, i.e., the upper threshold(noted as $u_u$) and 20\% the lower threshold(noted as $u_l$). The points with $T_{mag}$ that exceed the upper threshold are classified as frontal zone points, while below the lower threshold non frontal zone points and between the two thresholds undetermined frontal zone points. The CDF of one day is shown in figure \ref{fig:cdf}.

\subsection{Bayesian decision}
\begin{figure}
	\centering
	\includegraphics[width=0.1\textwidth]{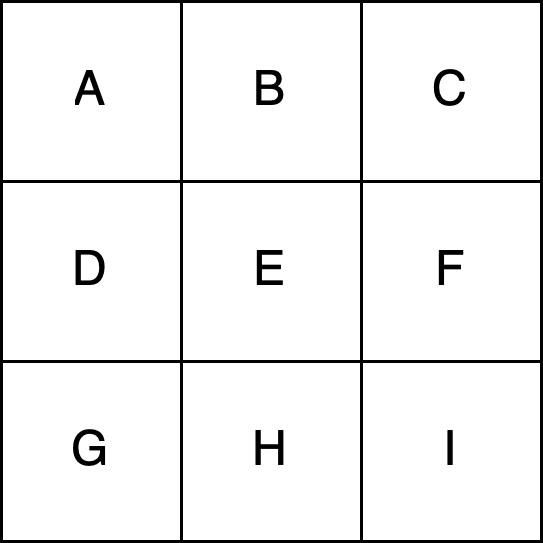}
	\caption{Undetermined frontal zone point} 
	\label{fig:surround} 
\end{figure}

For the undetermined frontal zone points obtained above, a Bayesian decision procedure is imposed to judge if the points are frontal zone points. As shown in figure \ref{fig:surround}, suppose the point to be determined is $E$, which is surrounded by points $A, B, C, D, F, G, H$ and $I$.

\begin{equation}\label{eq:prior}
	\begin{gathered}
		P(E=i)=\frac{T_E-u_l}{u_u-u_l}\\
		P(E=j)=\frac{u_u-T_E}{u_u-u_l}
	\end{gathered}
\end{equation}

Firstly, the prior probability of $E$ being frontal zone point or non frontal zone point is calculated with \eqref{eq:prior} where $i$ is frontal zone point, $j$ is non frontal zone point and $T_E$ is gradient magnitude of $E$.

It is clear that the prior probability is a linear transform of gradient magnitude and they are positively relevant. In another word, prior probability is the information of first derivative.

Secondly, likelihood is obtained by using more information of the whole field, i.e., local degree of edge(LDE) and block deviation(BD)\cite{Ping-2014}. LDE and BD are calculated for the $4$ pairs of points($A-I$, $B-H$, $C-G$ and $D-F$) around $E$. 

\begin{equation}\label{eq:lde_bd_ai}
	\begin{gathered}
		LDE(A,I)=\frac{4}{7}\frac{V_{max}-\overline{V}-|A-I|}{V_{max}-V_{min}}+\frac{1}{2}\\
		BD(A,I)=\frac{|A-I|}{V_{max}-V_{min}}
	\end{gathered}
\end{equation}

In \eqref{eq:lde_bd_ai}, $V=[A, B, C, D, F, G, H, I]$ is the vector of $8$ SST values around $E$. $V_{max}$ is the maximum of these components in $V$, $V_{min}$ is the minimum and $\overline{V}$ the average. It can be verified that LDE and BD are two mathematical operators which fall into $[0, 1]$ for normalization\cite{Mansoori-2006, Bauer-1996}. 

It can also be concluded that $LDE(A,I)$ is the similarity between $|A-I|$ and local maximum amplitude($V_{max}-\overline{V}$) and $BD(A,I)$ is normalized gradient magnitude in $A-I$ direction.

\begin{equation}\label{eq:lde_bd_e}
	\begin{gathered}
		LDE(E)=\frac{1}{4}(LDE(A,I)+LDE(B,H)+LDE(C,G)+\\
		LDE(D,F))\\
		BD(E)=\frac{1}{4}(BD(A,I)+BD(B,H)+BD(C,G)+\\
		BD(D,F))\\
	\end{gathered}
\end{equation}

The process is the same for $B-H$, $C-G$ and $D-F$. LDE and BD average of these $4$ pairs generate the LDE and BD of $E$, as shown in \eqref{eq:lde_bd_e}.

\begin{equation}\label{eq:likelihood}
	\begin{gathered}
		P(fact|E=i)=\frac{F_{LDE,E}}{F_{E}}\frac{F_{BD,E}}{F_{E}}\\
		P(fact|E=j)=\frac{NF_{LDE,E}}{NF_{E}}\frac{NF_{BD,E}}{NF_{E}}
	\end{gathered}
\end{equation}

Then comes the calculation of likelihood. In \eqref{eq:likelihood}, $F_E$ represents the number of points whose gradient magnitude is greater than $T_E$, $F_{LDE,E}$ is the number with LDE difference smaller than $0.1$ in $F_E$ compared to $LDE(E)$ and $F_{BD,E}$ is the same logic for BD difference. The derivation of $NF_E$, $NF_{LDE,E}$ and $NF_{BD,E}$ follows the same steps as described above.

The term $fact$ in \eqref{eq:likelihood} is based on the assumption that if $E$ is a frontal zone point($E=i$), then LDE and BD of greater gradient magnitude points should be similar and vice versa. The assumption is natural for it introduces $2$ mathematical operators, i.e. patterns to generate likelihood.

\begin{equation}\label{eq:bayesian}
	\begin{gathered}
		P(E=i|fact)=\frac{P(fact|E=i)P(E=i)}{P(fact)}\\
		P(E=j|fact)=\frac{P(fact|E=j)P(E=j)}{P(fact)}
	\end{gathered}
\end{equation}

\begin{equation}\label{eq:decision}
	{E} = \begin{cases}
		i,&{\text{if}}P(fact|E=i)P(E=i) \geq P(fact|E=j)\\
		&P(E=j)\\
		{j,}&{\text{otherwise.}} 
	\end{cases}
\end{equation}

\begin{algorithm}
	\caption{Bayesian decision.}
	\label{alg:bd}
	\begin{algorithmic}[1]
		\Require
		SST and SST gradient magnitude;
		\Ensure
		Frontal zone;
		\State Use SST gradient magnitude to get prior probability for every point by \eqref{eq:prior};
		\label{code:fram:prior}
		\State Use SST to calculate LDE and BD by \eqref{eq:lde_bd_ai} and \eqref{eq:lde_bd_e};
		\label{code:fram:lde_bd}
		\State Obtain likelihood by \eqref{eq:likelihood}; 
		\label{code:fram:likelihood}
		\State Get posterior and make Bayesian decision by \eqref{eq:bayesian};
		\label{code:fram:posterior}
		\Return Frontal zone;
	\end{algorithmic}
\end{algorithm}

Finally, with Bayes theorem and decision for this paper--\eqref{eq:bayesian}\eqref{eq:decision}, the undetermined frontal zone point $E$ can be judged by posterior probability. If $P(fact|E=i)P(E=i)>=P(fact|E=j)P(E=j)$, then $E$ is seen as frontal zone point and non frontal zone point otherwise. The whole process is shown in algorithm \ref{alg:bd}.

\subsection{Skeletonization}
With maximum disk method(MDM) in mathematical morphology, these one-pixel-width fronts can be extracted from the frontal zone obtained above. MDM is a skeletonization algorithm that determines whether each point in the frontal zone is a front point, also known as a skeleton point. The specific principle is to expand the disk with each point as the center. When there are two or more tangent points with the boundary during the expansion process, it is determined as a skeleton point. 

\begin{figure}
	\centering
	\includegraphics[width=0.15\textwidth]{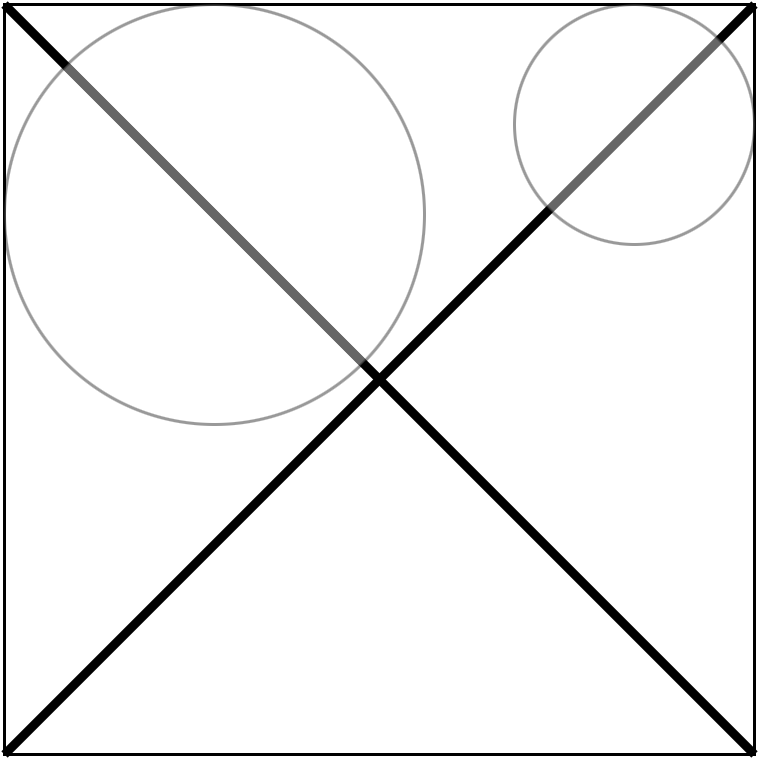}
	\caption{Examples of MDM} 
	\label{fig:mdm} 
\end{figure}

\begin{equation}\label{eq:mdm}
	\begin{gathered}
		S(Z)=\bigcup_{k=0}^K S_k (Z)\\
		S_k (Z)=(Z \ominus kM)-(Z \ominus k)\degree M\\
		K= \max \{k|(Z \ominus kM) \neq \varnothing \}
	\end{gathered}
\end{equation}

The principle can be seen in figure \ref{fig:mdm}, and the mathematical principle can be seen in \eqref{eq:mdm}. In it, $Z$ is the determined frontal zone, $(Z \ominus kM)$ means performing $k$ rounds of corrosion on the identified frontal zone, $M$ is the convolutional kernel, $\degree$ is the open operation and $K$ is the maximum number of corrosion cycles before being corroded into an empty set.

\subsection{Skeleton trimming}
Although these fronts have been found, they have many branches, and in order to extract the main features and adapt to the research scale, redundant branches must be removed. Here, a greedy algorithm called the discrete skeleton evolution(DSE) algorithm is used to assign weights to each branch, iteratively removing the branch with the minimum weight to obtain the main fronts\cite{Bai-2007, Vincent-1993}.

The points in $S(Z)$ need to be classified into $3$ categories: endpoint, connection point and intersection point. The endpoint has $1$ point in $S(Z)$ connected to itself, the connection point has $2$ and the intersection point has $3$. It is clear to see that endpoint and intersection point determine a branch in $S(Z)$. Suppose $l_i(i=1,2,...,N)$ are the endpoints in $S(Z)$, $f(l_i)$ are the closest intersection points to $l_i$, $Q(l_i,f(l_i))$ are the branches.

The frontal zone after skeleton reconstruction was determined according to $R(S)=\cup_{s\in S(Z)}U(s,r(s))$; among them, $R(S)$ is the frontal zone after skeleton reconstruction, $r(s)$ is the radius of the largest disk $U(s,r(s))$ with the center $s$ and in the frontal zone.

\begin{equation}\label{eq:weight}
	w_i=V(R(S))-V(R(S-Q(l_i,f(l_i))))
\end{equation}

Using \eqref{eq:weight} to assign weights to every branch $Q(l_i,f(l_i))$ and $V()$ is the area function. $w_i$ is the number of pixels loss when comparing the reconstructed frontal zone and the original. It can be seen from this equation that if $w_i$ approches $0$, it will mean that the reconstructed frontal zone after removing $Q(l_i,f(l_i))$ is similar to the original, so it can be removed and a greedy algorithm can be obtained as in algorithm \ref{alg:dse}.

\begin{algorithm}
	\caption{DSE.}
	\label{alg:dse}
	\begin{algorithmic}[1]
		\Require
		The original skeleton of frontal zone $S^0(Z)$(S(Z));
		\Ensure
		Trimmed skeleton $S^{K}(Z)$(assume $K$ iterations);
		\State Set an upper threhold $t$ for minumum weight;
		\label{code:fram:threshold}
		\State In the $kth$ iteration, update weights $w^k_i$ in $S^k(Z)$ according to  \eqref{eq:weight};
		\label{code:fram:kth_weight}
		\State Choose the mimimum weight $w^k_{min}$; 
		if $w^k_{min} \leq t$, skip to \ref{code:fram:dse_trim} otherwise \Return $S^{k}(Z)$;
		\label{code:fram:dse_judge}
		\State Remove the branch and update skeleton: $S^{k+1}(Z)=S^k(Z)-Q(l_{min}^k,f(l_{min}^k))$;
		skip to \ref{code:fram:kth_weight};
		\label{code:fram:dse_trim}
	\end{algorithmic}
\end{algorithm}

\begin{figure}
	\centering
	\includegraphics[width=0.45\textwidth]{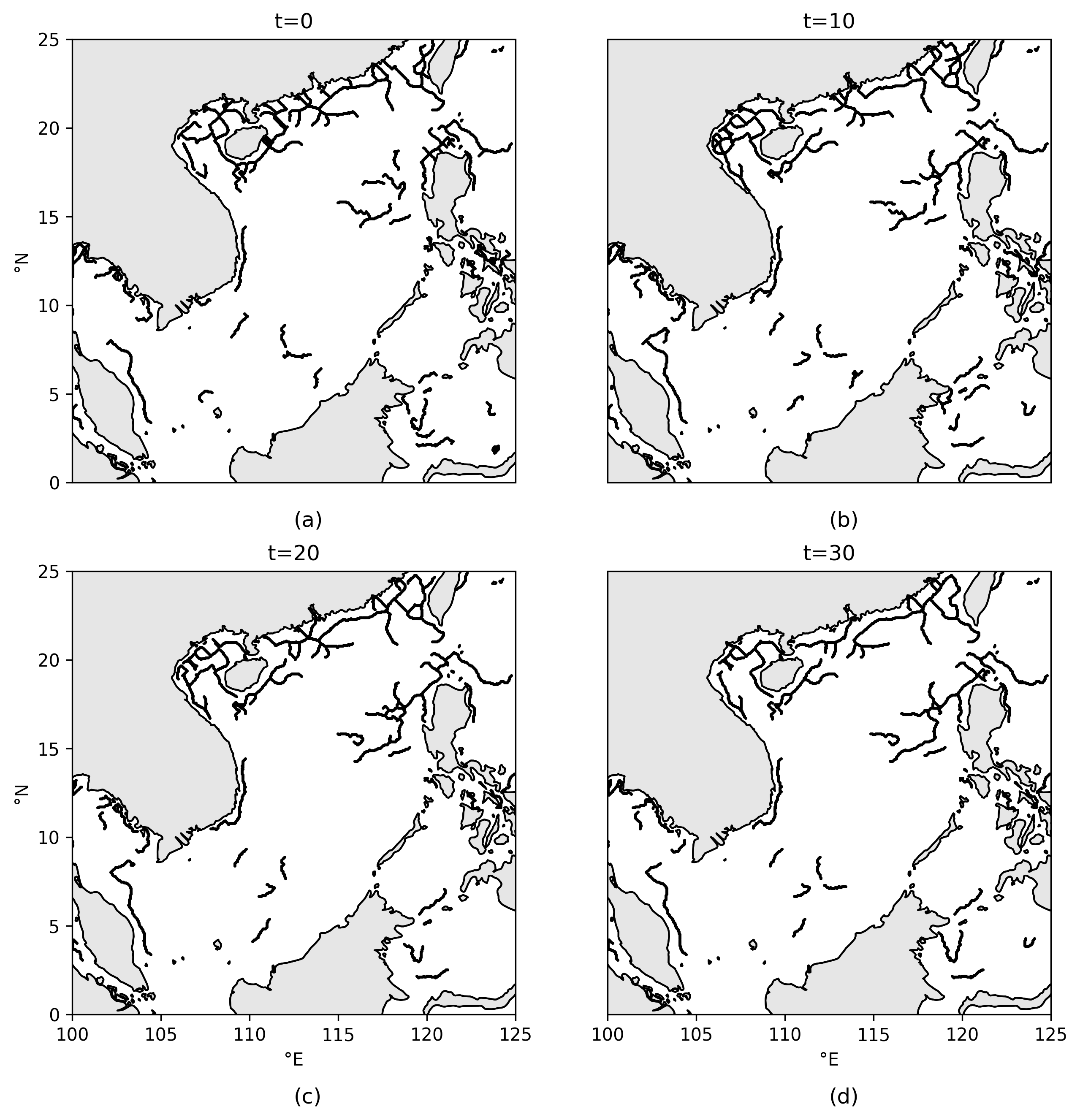}
	\caption{SST fronts in 2025.01.01 with different $t$} 
	\label{fig:dse} 
\end{figure}

Figure \ref{fig:dse} is a demo of DSE and it shows the influence of threshold $t$ on the performance for SST fronts in Jan 1st , 2025 in SCS.

\section{Methodology}
\subsection{Following, merging and filling}
For the one-pixel-width fronts obtained above, the commonly used DFS algorithm is applied to count every front and vector data is stored. The data structure of fronts in one day is in the form of nested linked list: $[[(Lon_1,Lat_1), (Lon_2,Lat_2),\ldots,(Lon_N, Lat_N)], [], []]$. Here, every component in the outer linked list is also a linked list(front), it is composed of theses front points' latitude and longitude.

As to front merging problem, this paper focuses on $2$ primary principles: (1)distance; (2)gradient direction and created a diagonal-direction-first filling method to fill in the gaps between nearby fronts.

\begin{figure}
	\centering
	\subfloat[Front merging and filling \label{fig:merge}]{\includegraphics[width=0.2\textwidth]{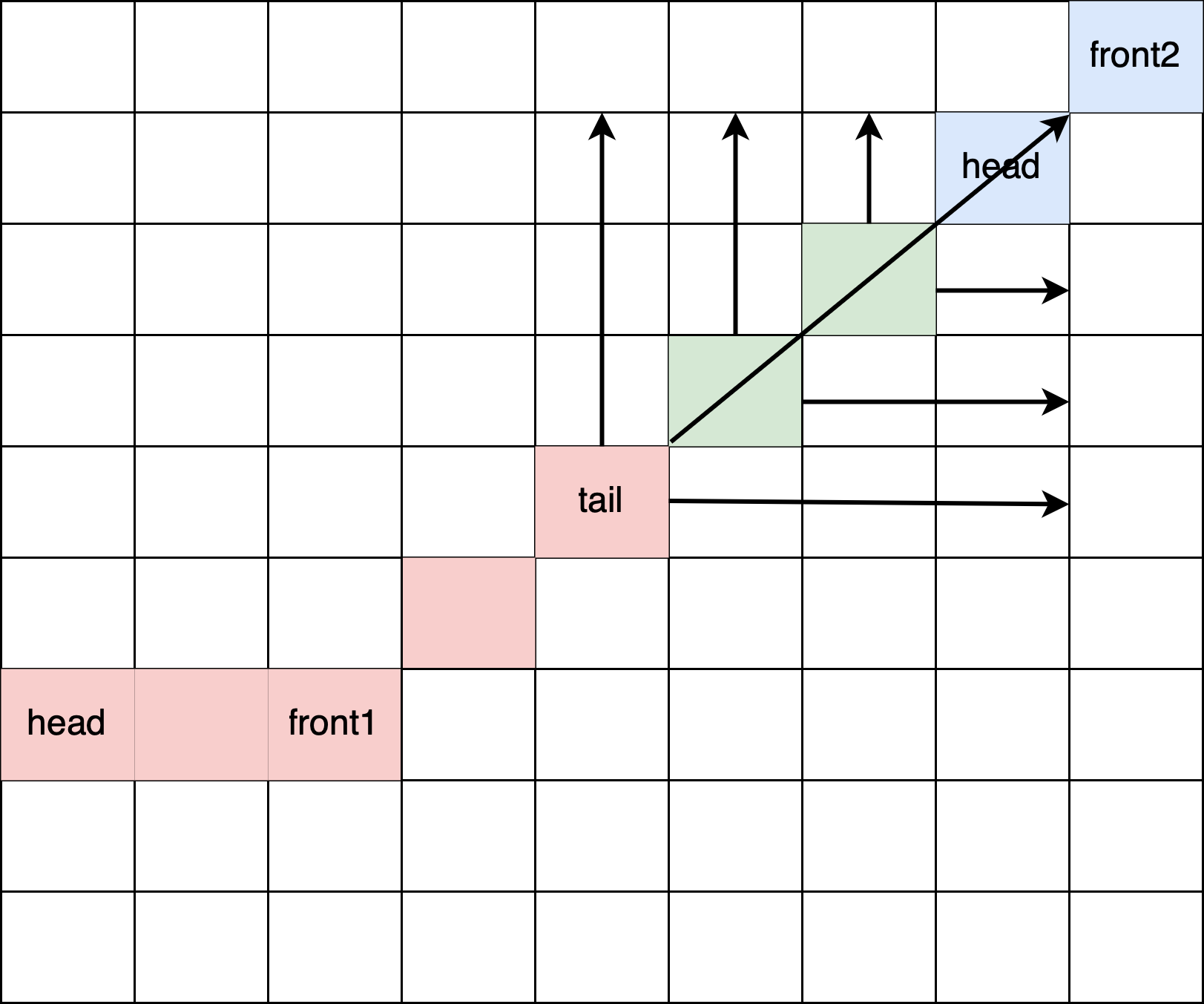}} \hspace{0.05\textwidth}
	\subfloat[Ring deletion \label{fig:ring}]{\includegraphics[width=0.2\textwidth]{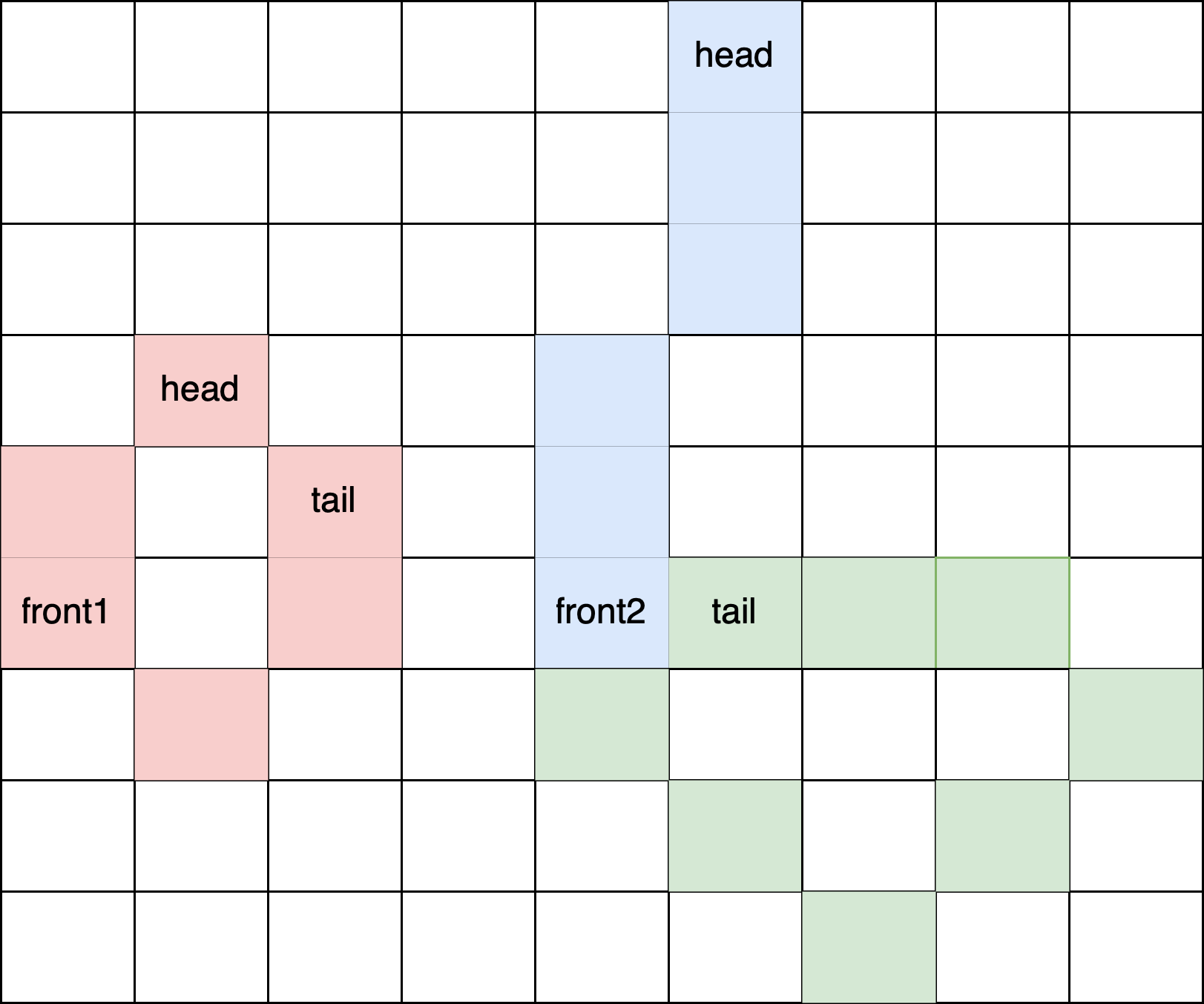}}
	\caption{Display of $2$ algorithms}
\end{figure}

\begin{algorithm}
	\caption{Merging and filling.}
	\label{alg:merge}
	\begin{algorithmic}[1]
		\Require
		Front vector data $F=[F_1,...,F_N]$;
		\Ensure
		Front vector data $F=[F_1,...,F_M]$;
		\State Set a search radius;
		\label{code:fram:radius}
		\State For one front $F_p$, if exists another front's head/tail in the search radius of its head/tail, put it on the waiting list $L_p$; if $L_p \neq \varnothing$, skip to \ref{code:fram:determine}, otherwise change another front $F_q$ and restart \ref{code:fram:search}; if all of the waiting list is empty, \Return $F$;
		\label{code:fram:search}
		\State For fronts on the waiting list, choose the nearest front to merge; if more than $2$ fronts are nearest, choose the least gradient direction difference to merge;
		\label{code:fram:determine}
		\State Merge and fill according to diagonal-direction-first principle; update $F$ and skip to \ref{code:fram:search};
		\label{code:fram:fill}
	\end{algorithmic}
\end{algorithm}

This algorithm can be illustrated by algorithm \ref{alg:merge} and figure \ref{fig:merge}. The green points is the filled points between $2$ nearby fronts and filling is in the arrow direction.

\subsection{Ring deletion}
As to the fronts obtained above, there exists ring structure. It is usually generated by eddies and is not the object for study in this paper, so a ring deletion algorithm is proposed.

\begin{algorithm}
	\caption{ Ring deletion.}
	\label{alg:ring}
	\begin{algorithmic}[1]
		\Require
		Front vector data $F=[F_1,...,F_M]$;
		\Ensure
		Front vector data $F=[F_1,...,F_L]$;
		\State if $F$ is changed, skip to \ref{code:fram:whole_ring}, otherwise \Return $F$;
		\label{code:fram:judge_same}
		\State For every $F_p$ in $F$, if its head is in its tail's $3\times 3$ nearby region, then $F_q$ is a ring structure itself, delete $F_p$ from $F$ and skip to \ref{code:fram:judge_same}; if no component in $F$ is ring structure itself, skip to \ref{code:fram:partial_ring}
		\label{code:fram:whole_ring}
		\State For every $F_p$ in $F$, from its head judge if any point is in its tail's $3\times 3$ nearby region; if exists, delete this part from $F_p$ and skip to \ref{code:fram:judge_same};
		\label{code:fram:partial_ring}
	\end{algorithmic}
\end{algorithm}

Algorithm \ref{alg:ring} illustrates the logic of ring deletion. In figure \ref{fig:ring}, front1 is a ring structure itself and front2 comprises a partial ring structure--the green blocks.

\subsection{Front tracking algorithm}
\subsubsection{Length ratio decision}
\begin{equation}\label{eq:length_decision}
	\frac{\min(N, M)}{\max(N, M)} < 0.5
\end{equation}

Consider fronts in $2$ consecutive days and assume the former day has $P$ fronts, the latter day $Q$. Assume $F_1=[(Lon_1, Lat_1), (Lon_2, Lat_2), ..., (Lon_N, Lat_N)]$ is the $1st$ front in the former day and $f_1=[(lon_1, lat_1), (lon_2, lat_2), ..., (lon_M, lat_M)]$ is the $1st$ front in the latter day. It is natural to think that if the length difference of $F_1$ and $f_1$ is too much, they can not be considered as the same front even if they are very close, so a length ratio decision should be made according to \eqref{eq:length_decision}. Here, the ratio threshold is set at $0.5$, if the ratio exceeds $0.5$, $F_1$ and $f_1$ will be taken as the same front in different date, otherwise independent fronts.

\subsubsection{First and second distance}
\begin{figure}
	\centering
	\subfloat[First and second distance \label{fig:first_second}]{\includegraphics[width=0.35\textwidth]{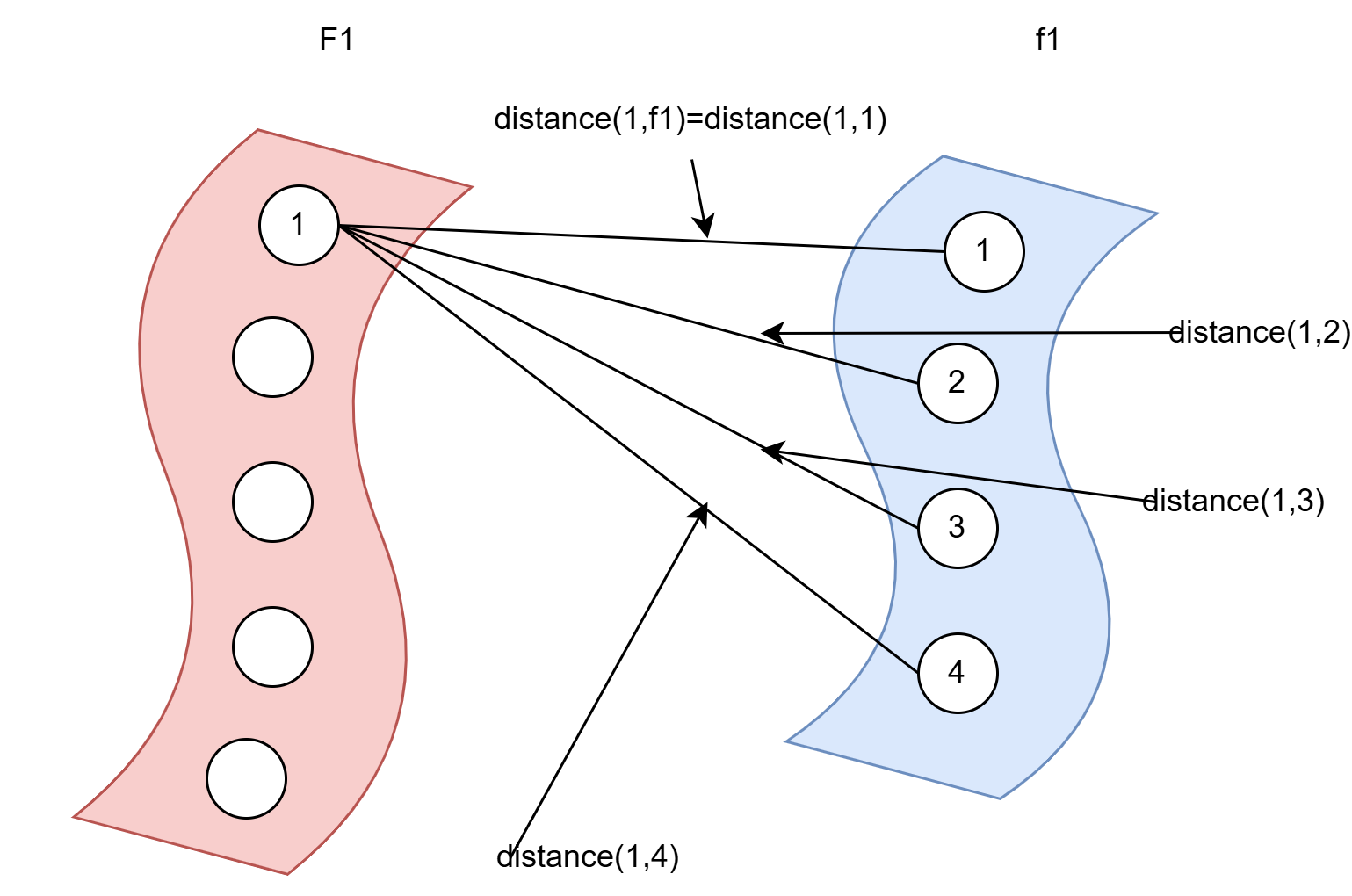}} \hspace{0.05\textwidth}
	\subfloat[Third distance \label{fig:third}]{\includegraphics[width=0.32\textwidth]{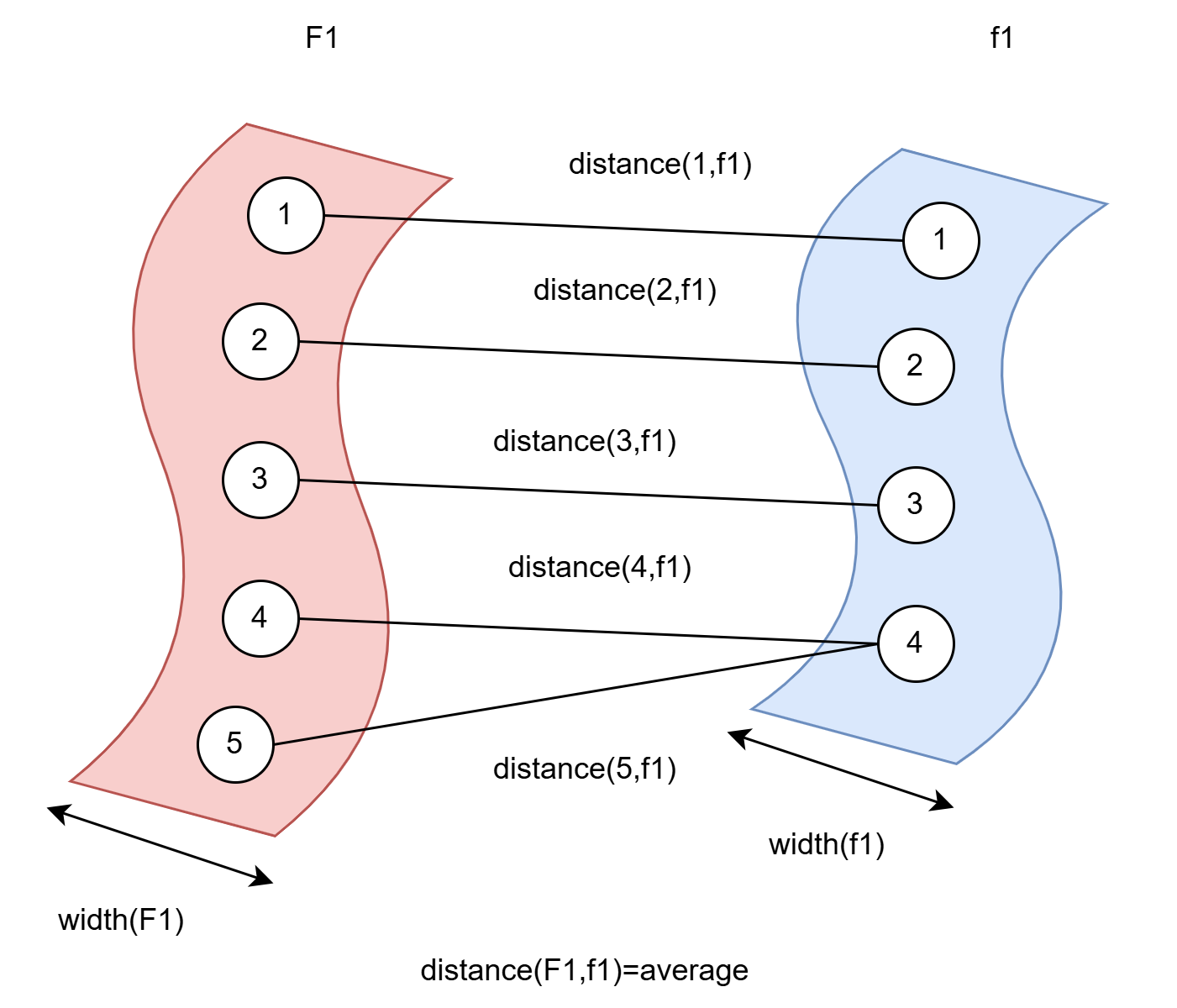}}
	\caption{Three distances}
\end{figure}

\begin{equation}\label{eq:first_distance}
	\begin{aligned}
		distance(1,i)=& \sqrt{(Lon_1 -  lon_i)^2 + (Lat_1 -  lat_i)^2 }, \\
		& 1 \leq i \leq M\\
	\end{aligned}
\end{equation}

\begin{equation}\label{eq:second_distance}
	distance(1,f_1)= \min\limits_{1 \leq i \leq M} distance(1, i)
\end{equation}

Let $M \leq N$, the first distance is defined as the distance of one point in $F_1$ and one point in  $f_1$, as shown in \eqref{eq:first_distance}. The second distance is obtained by minimizing the first distance, it is the distance of one point in $F_1$ to $f_1$ in essence, as illustrated by \eqref{eq:second_distance} and figure \ref{fig:first_second}.

\subsubsection{Third distance}

\begin{equation}\label{eq:third_distance}
	distance(F_1, f_1) = \frac{\sum\limits_{j=1}^{N}distance(j, f_1)}{N}
\end{equation}	

With \eqref{eq:third_distance} and figure \ref{fig:third}, the third distance is defined as the average of the second distance, symbolizing the distance of $F_1$ and $f_1$. 

\subsubsection{Distance-width decision}
\begin{figure}
	\centering
	\includegraphics[width=0.3\textwidth]{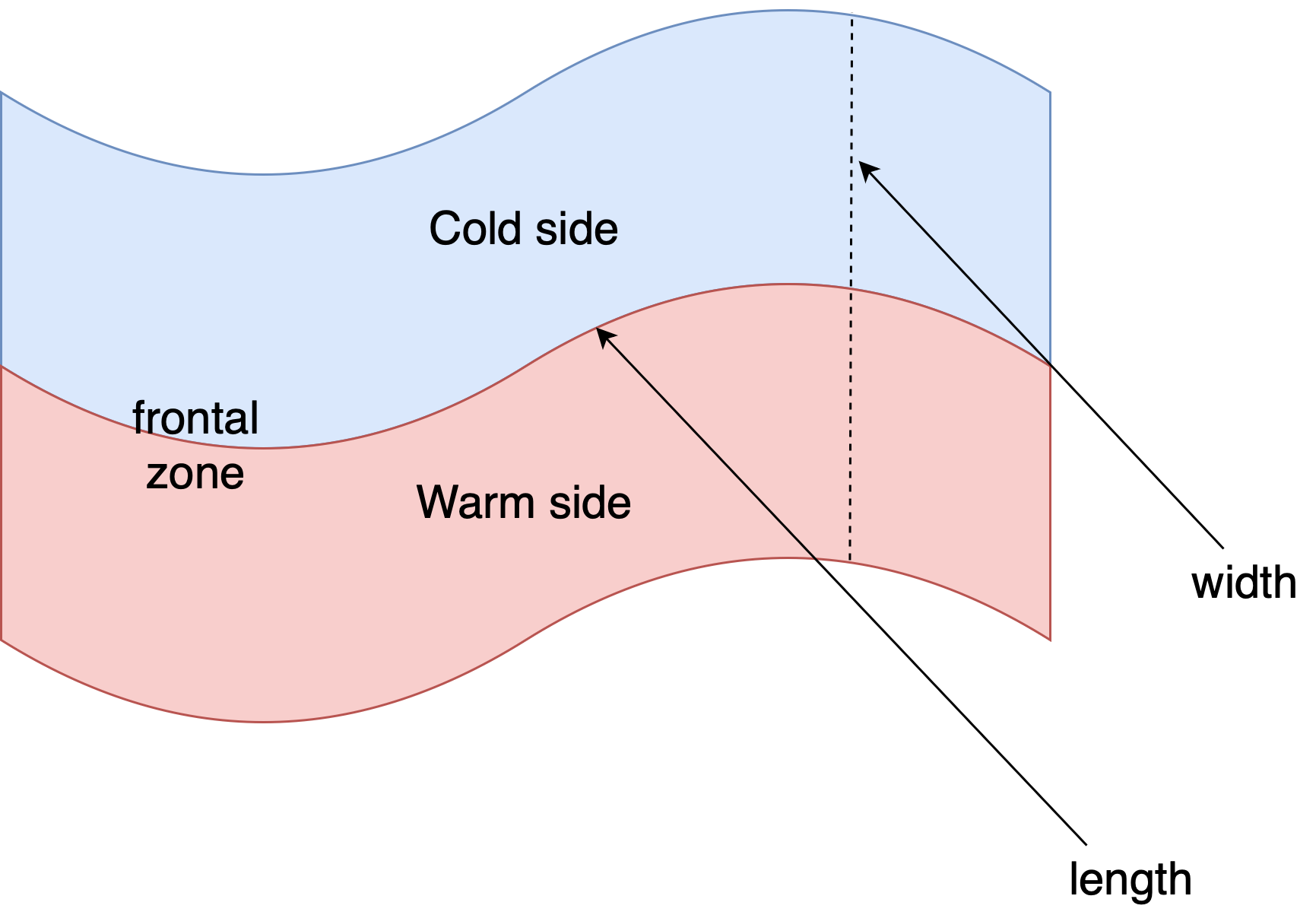}
	\caption{Front length and width} 
	\label{fig:width} 
\end{figure}

The width of each front is defined as the average value of twice the distance from the front point to the boundary of frontal zone, as shown in figure \ref{fig:width}.

\begin{equation}\label{eq:minimize}
	q = \mathop{argmin} \limits_{1 \leq k \leq Q}distance(F_1, f_k)
\end{equation}

The most similar front in the latter day to $F_1$ is obtained using \eqref{eq:minimize}.

\begin{equation}\label{eq:width_decision}
	 \begin{cases}
		{F_1=f_q},&{\text{if}} \, distance(F_1, f_q)\leq \frac{width(F_1)+width(f_q)}{2}\\
		{F_1 \neq f_q},&{\text{otherwise.}} 
	\end{cases}
\end{equation}

The decision process is shown in \eqref{eq:width_decision}. If the distance between $F_1$ and $f_q$ does not exceed their influence area, then it is judged as the same front and label them with the same id.

\begin{equation}\label{eq:metric_space}
	\begin{aligned}
		&Positivity: distance(\alpha, \beta) \geq 0, distance(\alpha, \beta) = 0 \Leftrightarrow \alpha=\beta\\
		&Symmetry: distance(\alpha, \beta) = distance(\beta, \alpha)\\
		&Triangular \, inequality: distance(\alpha, \beta) + distance(\beta, \gamma) \\ 
		&\geq distance(\alpha, \gamma)
	\end{aligned}
\end{equation}

It is easy to verify that these $3$ distances defined above satisfy the $3$ features of distance in metric space by \eqref{eq:metric_space}, so they are well-defined and can be extended to other fields. Here, $\alpha$, $\beta$ and $\gamma$ can be a point or front.

\begin{algorithm}
	\caption{ Tracking.}
	\label{alg:track}
	\begin{algorithmic}[1]
		\Require
		Front vector data of two days: $F=[F_1,...,F_P]$ and $f=[f_1,...,f_Q]$;
		\Ensure
		Front vector data of two days: $F=[F_1,...,F_P]$ and $f=[f_1,...,f_Q]$ with ids;
		\State For $F_p$(with no id) in $F$ and $f_q$ in $f$, perform length ratio decision by equation \ref{eq:length_decision}; put all of the passed $f_q$(with no id) in the waiting list; if the waiting list is empty, change another $F_p$ and restart \ref{code:fram:length_decision}, otherwise skip to \ref{code:fram:calculation_distance}; if the waiting list of all of the $F_p$ is empty, \Return $F$ and $f$(now with ids);
		\label{code:fram:length_decision}
		\State Calculate the $distance(F_p, f_q)$ for all $f_q$ in the waiting list and select the most similar $f_q$ by \eqref{eq:first_distance}, \eqref{eq:second_distance}, \eqref{eq:third_distance} and \eqref{eq:minimize};
		\label{code:fram:calculation_distance}
		\State Do distance-width decision by equation \ref{eq:width_decision}, if $F_p$ and $f_q$ are the same front, label them with the same id, update $F$ and $f$, skip to \ref{code:fram:length_decision};
	\end{algorithmic}
\end{algorithm}

The tracking algorithm can be symbolized in algorithm \ref{alg:track}.

\begin{figure*}
	\centering
	\includegraphics[width=0.9\textwidth]{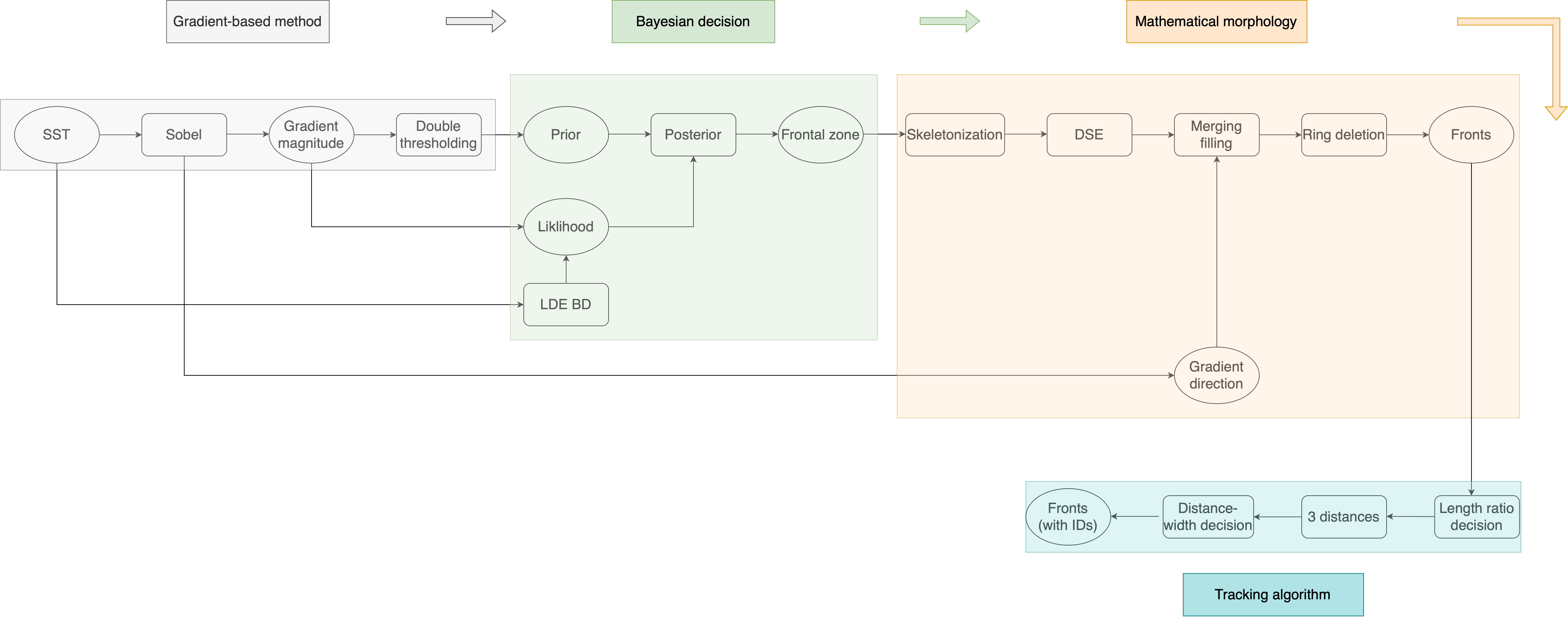}
	\caption{Flowchart of detection and tracking} 
	\label{fig:flowchart} 
\end{figure*}

The flowchart of front detection and trcaking can be seen in figure \ref{fig:flowchart}.

\section{Experiments}
\subsection{Case results}
\label{ch:case}
\subsubsection{Detection}
\begin{figure*}
	\centering
	\includegraphics[width=0.8\textwidth]{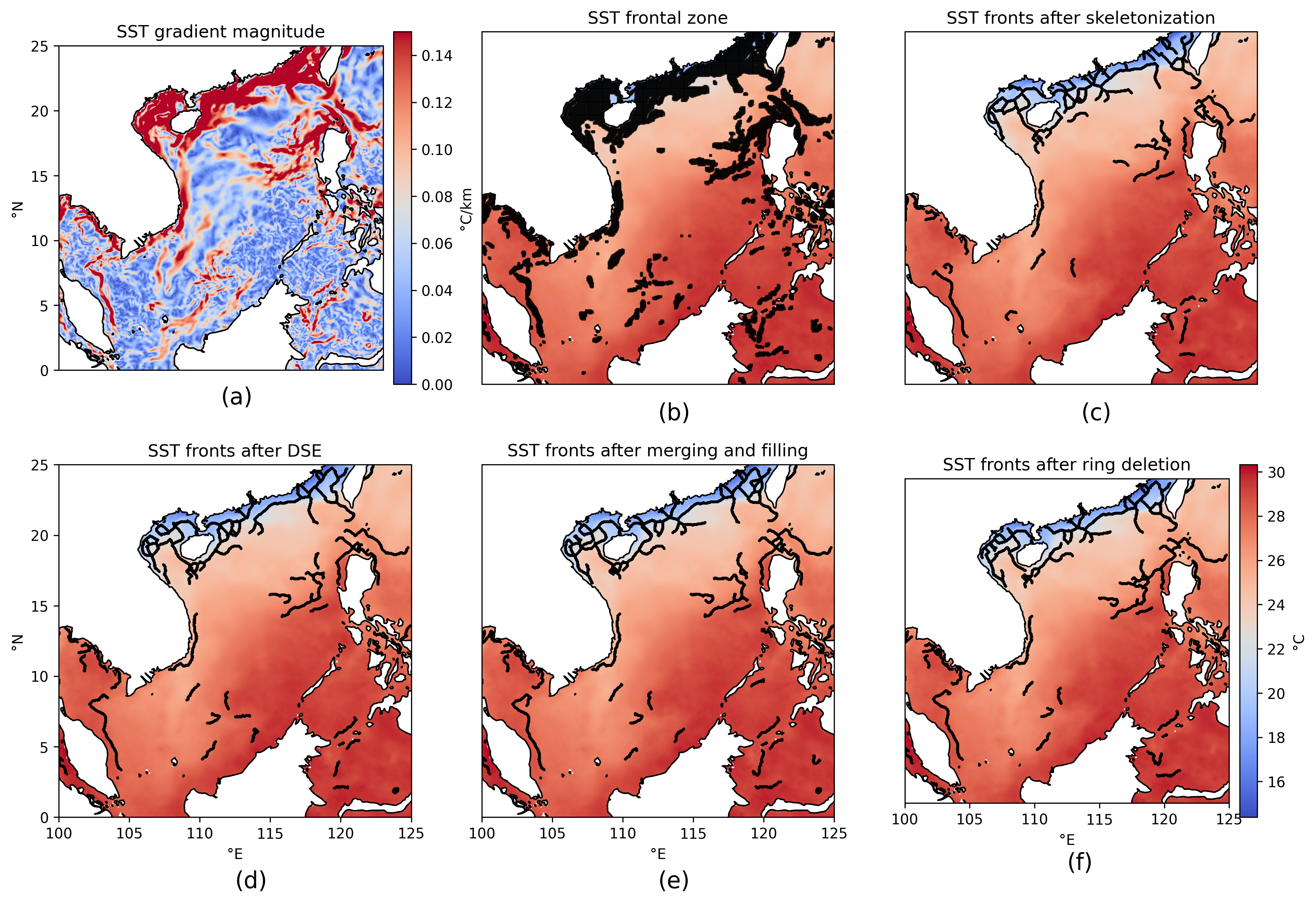}
	\caption{The detection result for SST in 2025.01.01} 
	\label{fig:case_study} 
\end{figure*}

Figure \ref{fig:case_study} shows the detection result of SCS SST in  Jan 1st, 2025. The research scale is set at $100$km, so in (c), the fronts shorter than $100$km($20$ pixels) are removed for this work focuses on large-scale fronts. In (d), the $t$ for DSE algorithm is set at $20$ as the reconstruction area is at least $1$-pixel-width. In (e), the merge radius is set at $3$ and (f) shows the detection result after ring deletion.

\subsubsection{Tracking}
\begin{figure*}
	\centering
	\includegraphics[width=0.8\textwidth]{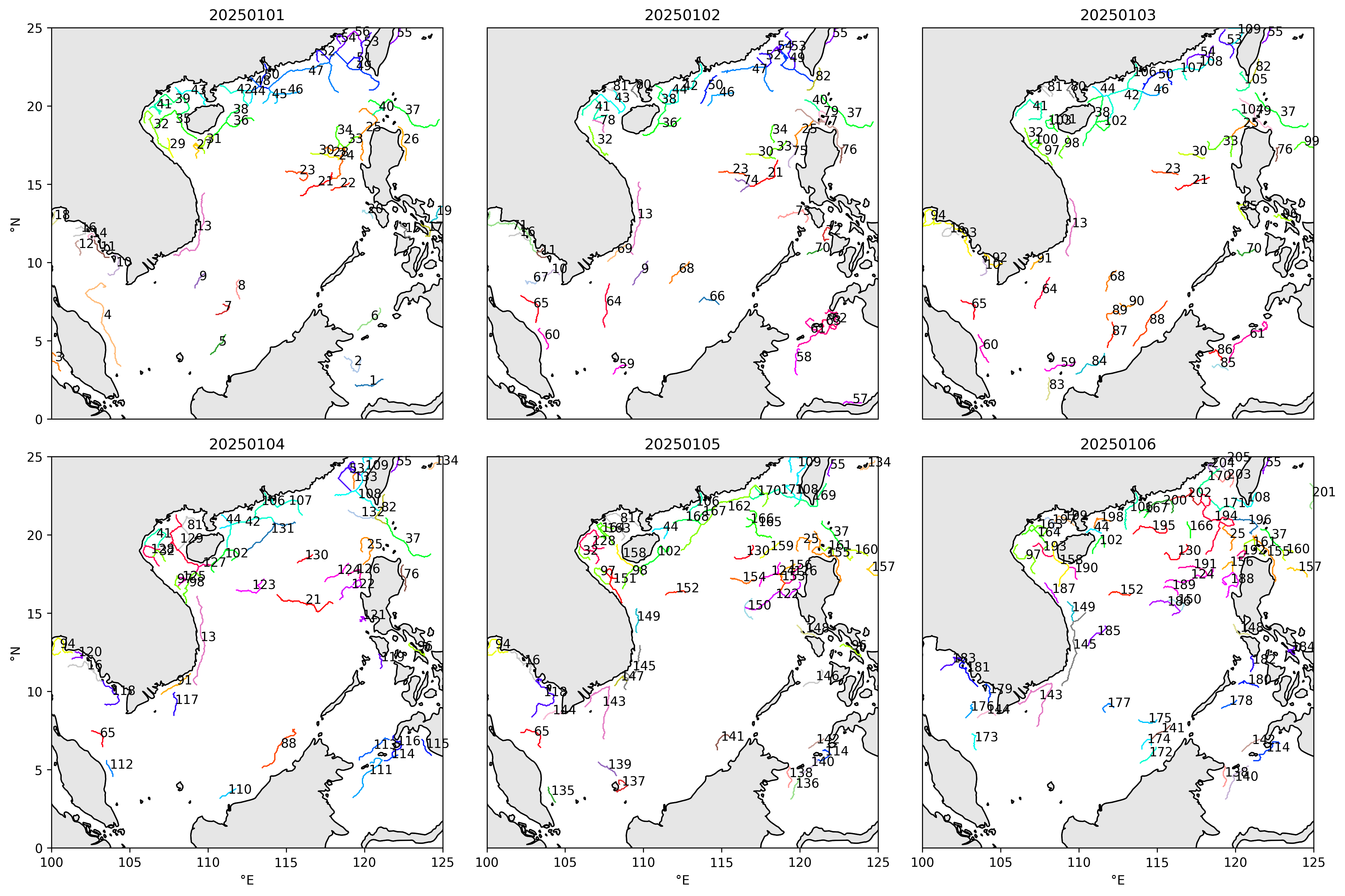}
	\caption{The tracking result for SST 2025.01.01-2025.01.06} 
	\label{fig:track} 
\end{figure*}

The tracking results for $6$ consecutive days in 2025 are shown in figure \ref{fig:track}. These fronts in different days are connected by the IDs on themselves.

\subsection{Statistical results}
\subsubsection{Detection}
\begin{figure}
	\centering
	\includegraphics[width=0.35\textwidth]{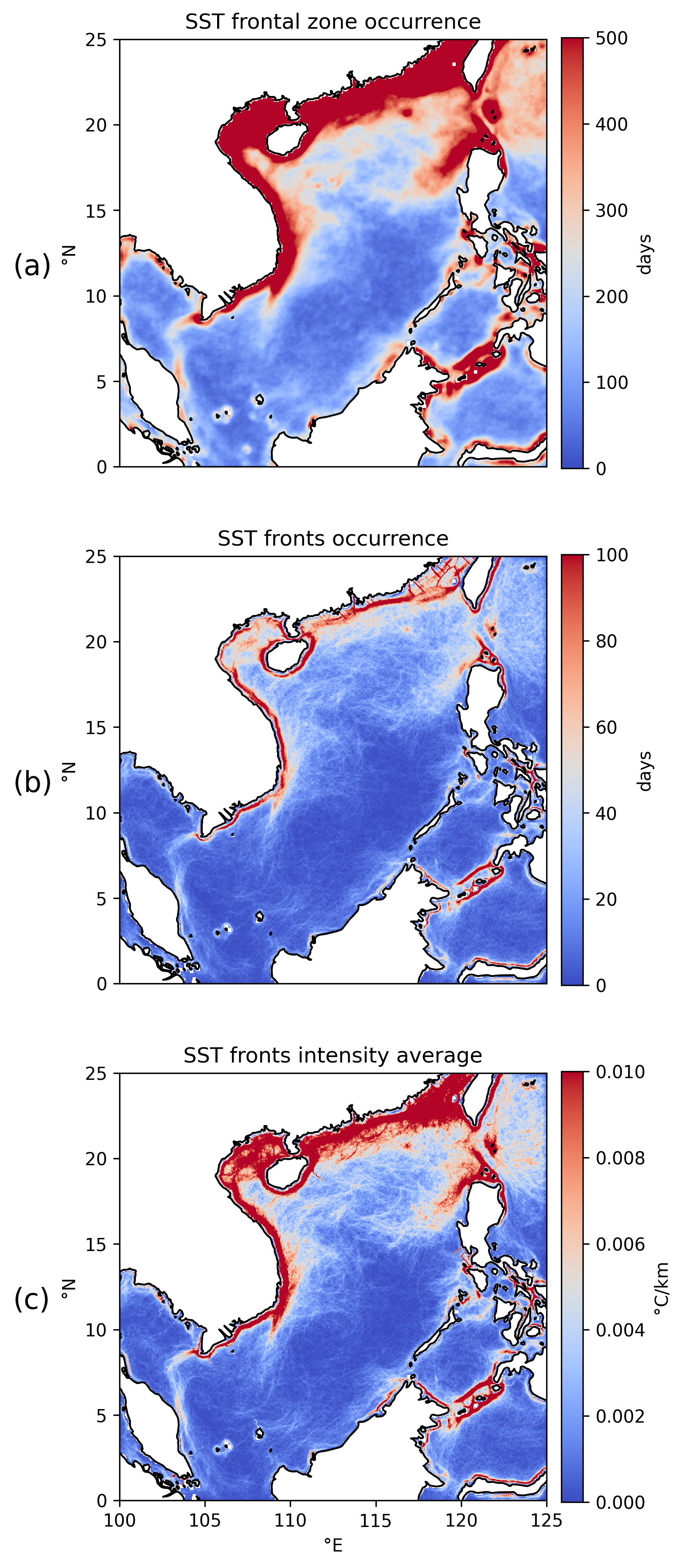}
	\caption{The detection result for SST from 2022 to 2024} 
	\label{fig:statistics} 
\end{figure}

The intensity of each front is defined as the average of gradient magnitude. Figure \ref{fig:statistics} (a-c) illustrates the frontal zone occurence, fronts occurence and average fronts intensity from 2022 to 2024. 

The higher occurrence of SST frontal zones in the northern parts of SCS, particularly in the latitudes between $10\degree$N and $20\degree$N, reflects seasonal transitions, where the differences between warm and cold ocean currents generate frontal zones. The seasonal shift in atmospheric circulation, coupled with oceanic upwelling and the influence of wind patterns, plays a crucial role in the formation of these frontal zones. The convergence of different water masses, such as warm tropical waters and cold subpolar currents, is responsible for the intensification of these frontal zones. 

Additionally, the northern and western parts of the map show the highest frequency of SST front occurrences, with areas along the eastern coast of SCS demonstrating more frequent front formations. This figure highlights higher SST front intensities along the eastern coast and in specific northern regions, particularly where strong temperature contrasts exist between adjacent water masses. The high-intensity areas are likely influenced by seasonal upwelling, where cold nutrient-rich waters rise to the surface, creating sharp contrasts with the warmer surface waters.

\begin{figure}
	\centering
	\includegraphics[width=0.35\textwidth]{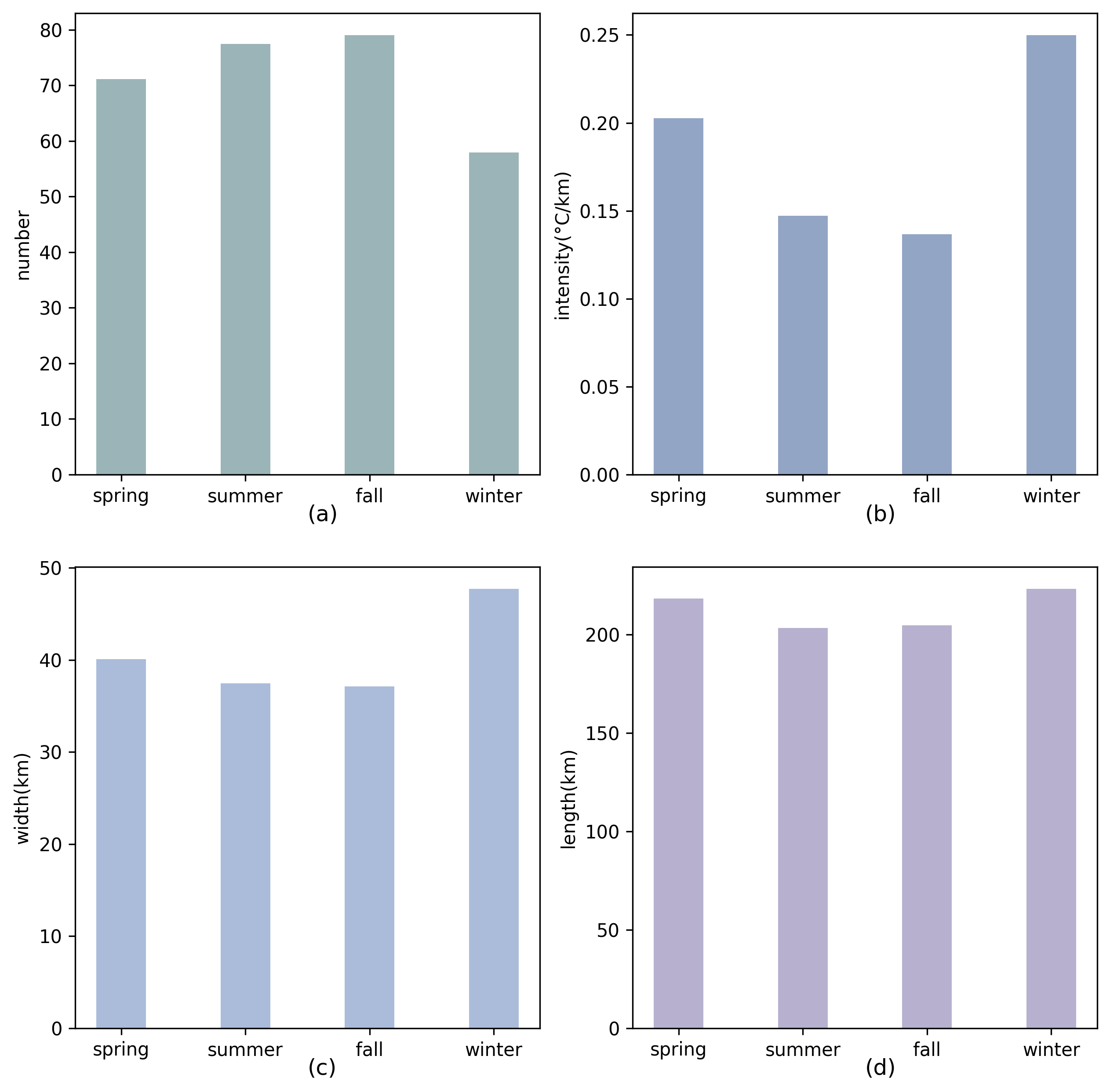}
	\caption{Seasonal average of SST front detection results from 2022 to 2024} 
	\label{fig:season} 
\end{figure}

Figure \ref{fig:season} displays the daily average detection result of four seasons from 2022 to 2024. Note that width and intensity are calculated weightedly according to length. Spring(March, April and May), summer(June, July and August), and fall(September, October and November) have similar numbers of ocean fronts, ranging between $71.14$ and $79.05$. Winter(December, January and February) shows a significant drop in the number of detected ocean fronts ($57.95$). Winter has fewer fronts due to lower atmospheric activity and less dynamic conditions, reducing the likelihood of ocean front formation.

Winter shows the highest intensity ($0.25\degree$C/km), followed by spring ($0.20\degree$C/km) summer ($0.15\degree$C/km) and fall ($0.14\degree$C/km). Intensity refers to the temperature gradient across the ocean front, which is typically stronger in spring and winter. Spring has more distinct temperature differences due to seasonal transitions, while winter's cold temperatures can create higher gradients. The lower intensity values in summer and fall indicate less temperature contrast between the ocean and surrounding atmosphere.

As to front width, winter has the widest ocean fronts ($47.72$km), followed by spring ($40.09$km) summer ($37.44$km) and fall ($37.11$km). Winter's colder temperatures and stronger winds lead to broader fronts, while spring's more dynamic conditions also create larger widths, but slightly smaller than winter.

Winter and spring have the longest ocean fronts, with lengths $223.29$km and $218.39$km, respectively. Fall and summer have shorter ocean fronts ($204.73$km and $203.38$km). It is due to atmospheric stability, lower storm activity and  high-pressure systems in winter and spring.

The fast fourier transform (FFT) is imposed on daily average of front number, intensity, width and length. The dominance of low-frequency components in the frequency spectra of all four features suggests that the primary driver of the variability in the number, intensity, width, and length of the fronts is likely associated with seasonal changes and long-term atmospheric patterns. 

\begin{figure*}
	\centering
	\includegraphics[width=0.8\textwidth]{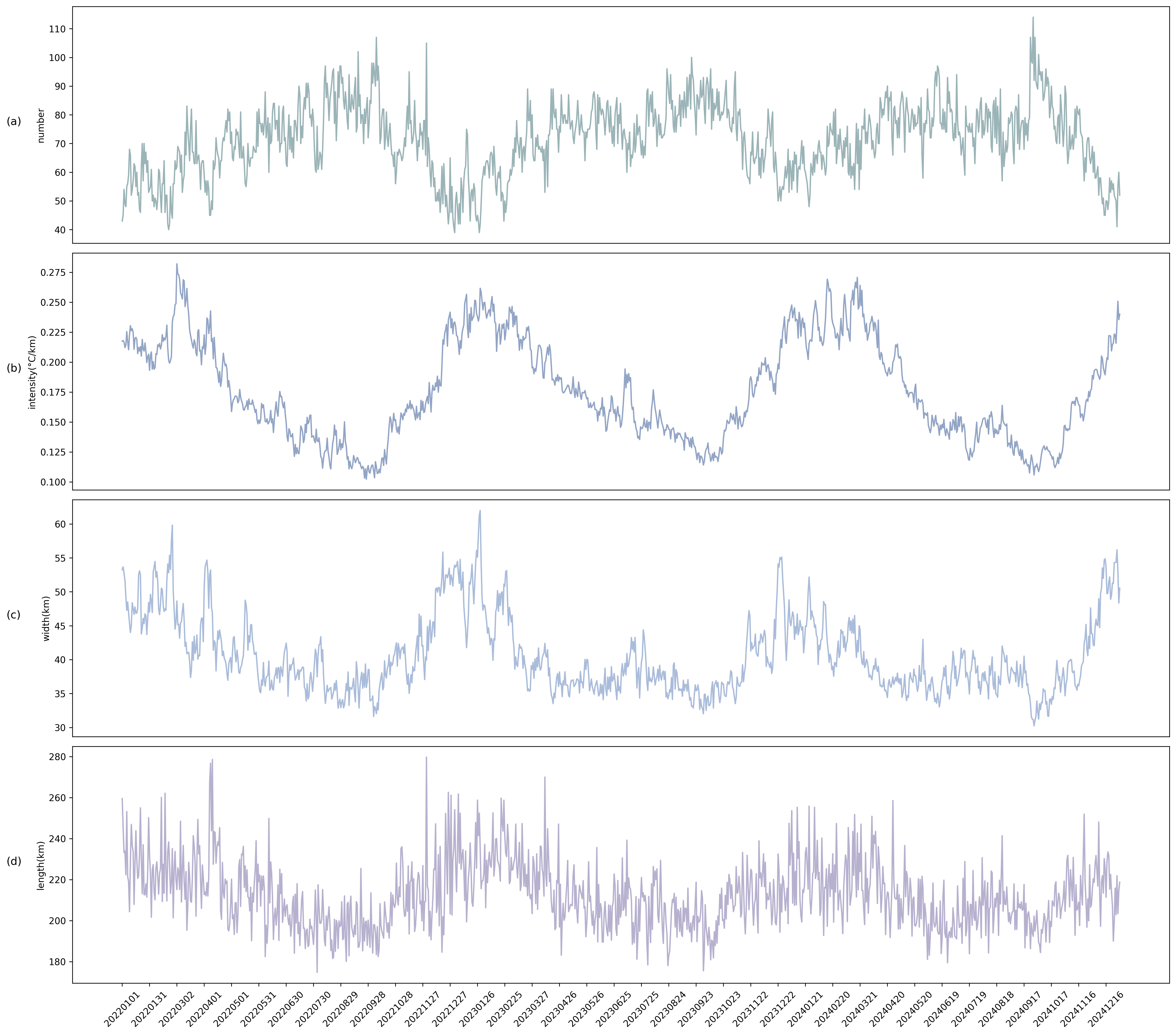}
	\caption{Daily average of SST front detection results from 2022 to 2024} 
	\label{fig:yearly_analysis} 
\end{figure*}

\begin{table}[]
	\caption{YEARLY AVERAGE OF SST FRONT DETECTION RESULTS}
	\label{tab:yearly}
	\centering
	\begin{tabular}{llll}
		\hline
		Feature/Year & 2022   & 2023   & 2024   \\ \hline
		Number       & 69.28  & 71.93  & 73.1   \\ 
		Intensity(℃/km)    & 0.17   & 0.18   & 0.18   \\
		Width(km)        & 41.84  & 40.11  & 39.75  \\
		Length(km)       & 212.69 & 213.03 & 211.54 \\ \hline
	\end{tabular}
\end{table}

\begin{figure*}
	\centering
	\includegraphics[width=0.8\textwidth]{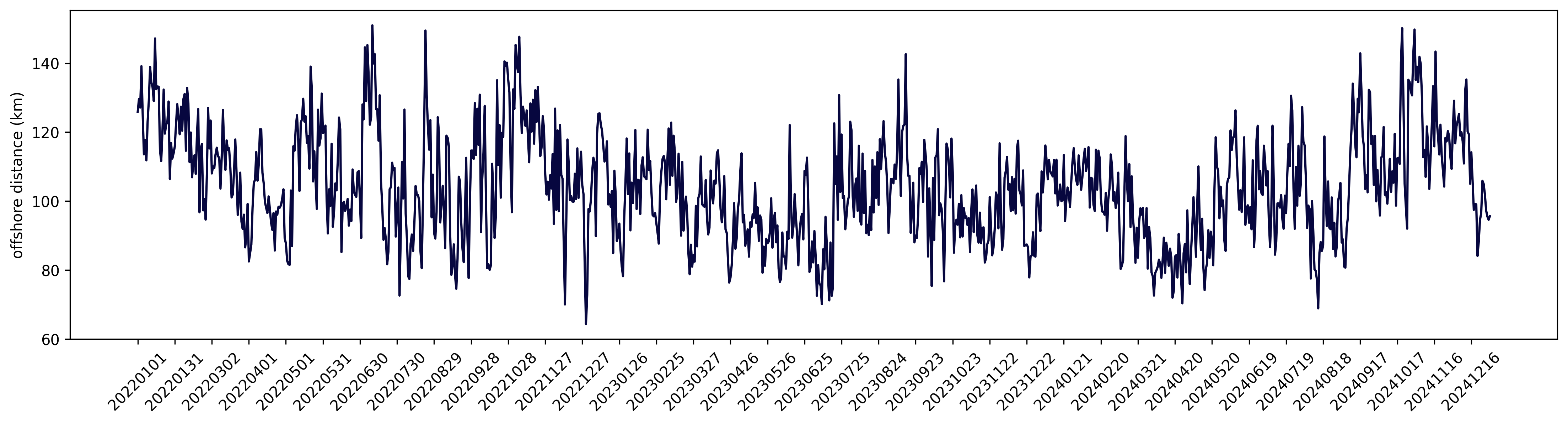}
	\caption{Daily average offshore distance of SST fronts from 2022 to 2024}
	\label{fig:offshore_dis}
\end{figure*}

\begin{figure}
	\centering
	\includegraphics[width=0.35\textwidth]{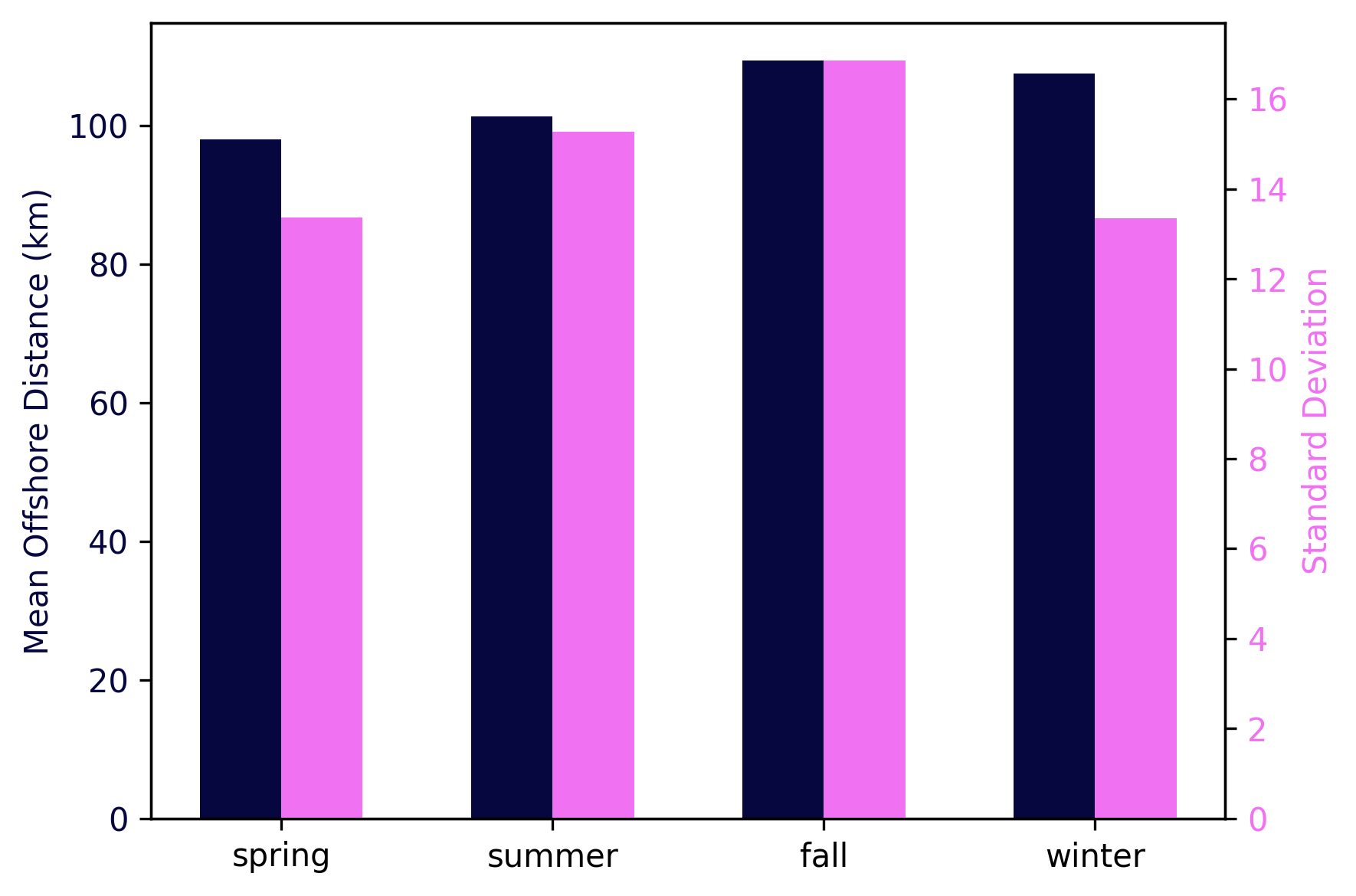}
	\caption{Seasonal mean and standard deviation of offshore distance from 2022 to 2024}
	\label{fig:offshore_mean_std}
\end{figure}

\begin{table}[]
	\caption{YEARLY AVERAGE OF OFFSHORE DISTANCE}
	\label{tab:yearly_offshore}
	\centering
	\begin{tabular}{llll}
		\hline
		Year & 2022   & 2023   & 2024   \\ \hline
		Distance(km)   & 109.56  & 98.88  & 103.53   \\ \hline
	\end{tabular}
\end{table}

The seasonal variation of SST front features can also be seen in figure \ref{fig:yearly_analysis}. Intensity, width and length follow the similar trend, while number is the adverse. Table \ref{tab:yearly} presents the yearly averages of SST front features. The silghtly increase in front numbers indicates a gradual rise in the frequency of front formations. There exists a small reduction in the spatial extent(width) of the fronts. Both of them are consistent with more frequent atmospheric disturbances over the past $3$ years. Intensity and length, however, remain relatively stable. 

Figure \ref{fig:offshore_dis} and \ref{fig:offshore_mean_std} show the daily and seasonal length-weighted average offshore distance of SST fronts respectively. The data exhibits pronounced seasonal oscillations, with offshore distance increasing during cooler seasons($109.30$km, $107.51$km for fall, winter)and decreasing in warmer seasons($97.99$km, $101.30$km for spring, summer). The standard deviation, however, differs greatly($13.37$, $15.27$, $16.85$ and $13.35$). The peak in fall is caused by more frequent weather changes, and the trough in winter and spring is attributed to relative stable weather conditions.

The drastic decline of offshore distance in 2023, as seen in table \ref{tab:yearly_offshore}, is consistent with the last El Nino phenomenon. During El Nino, the ocean's thermal structure is changed, causing fronts closer to the shore. In addition, the induced regional climatic anomalies such as storms and wind shifts can also push SST fronts closer to the shore.

\subsubsection{Tracking}
To test the performance of tracking, this paper uses intersection over union(IoU) of $2$ adjacent days' front IDs as an indicator. If IoU is close to $1$, then the fronts are not changed too much, otherwise there exists more fronts' position change, birth and death. 

\begin{figure}
	\centering
	\includegraphics[width=0.35\textwidth]{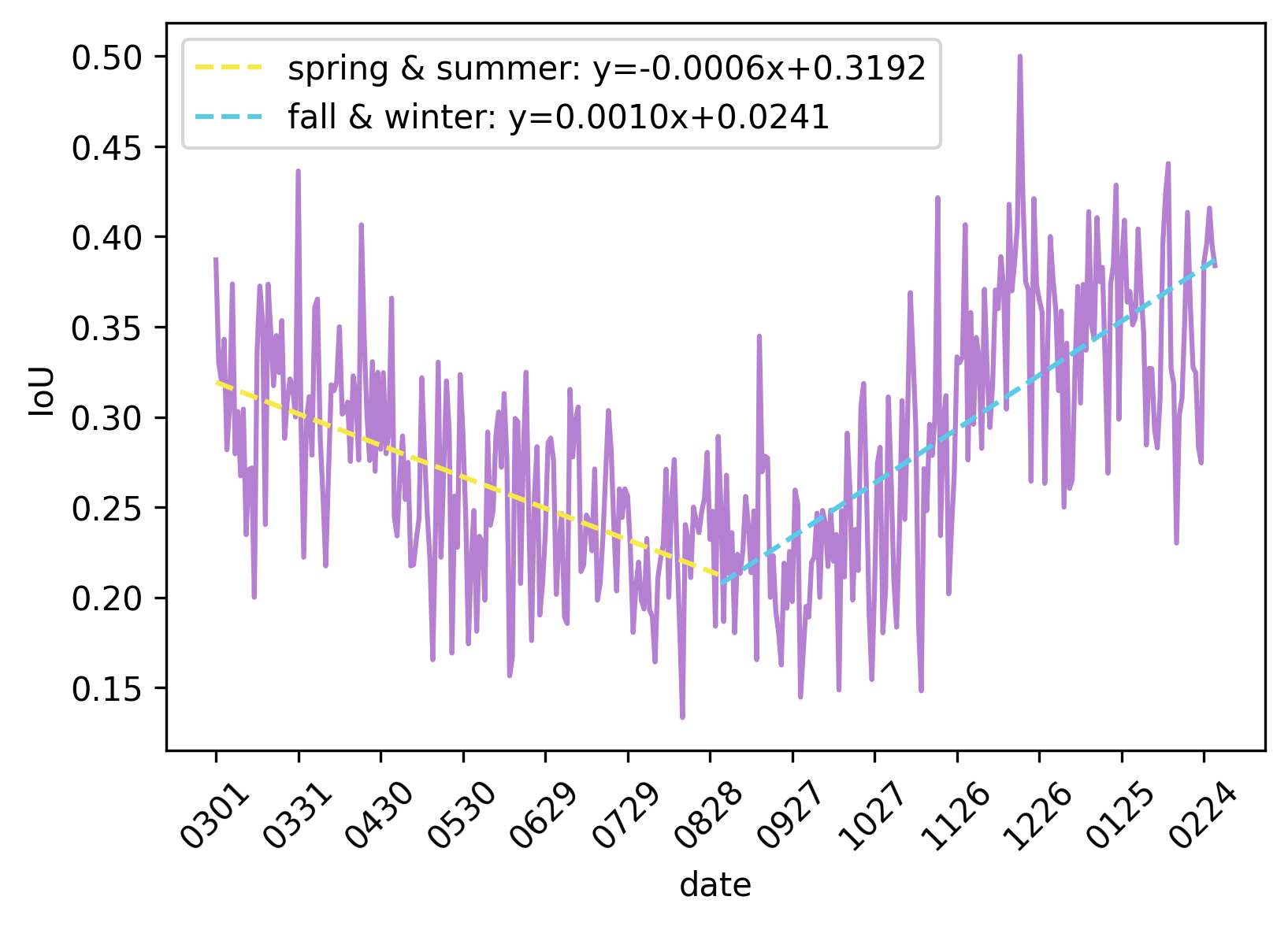}
	\caption{Iou of SST front tracking results in 2024} 
	\label{fig:iou} 
\end{figure}

Figure \ref{fig:iou} shows IoU values between two adjacent days' front IDs over a year, with the data segmented into different seasons (spring \& summer, fall \& winter) and their linear regression. The highest value is $0.5$, which occurs at 19th, Dec while lowest $1.33$ at 18th, Aug. The spring \& summer trend shows a negative slope, meaning that during that time, the IoU decreases slightly over time. The change in front positions, birth, and death events are more frequent in warmer months due to increased atmospheric instability, larger temperature contrasts, and stronger weather systems such as cyclones or thunderstorms. The trend in fall \& winter, however, shows a positive slope. This are attributed to the more stable atmospheric conditions often present in these colder months, with fewer atmospheric disturbances like storms and cyclones.

The higher frequency of weather changes during the spring and summer months, caused by stronger temperature gradients, increased humidity, and storm systems, leads to more birth and death of fronts. The variability in front positions is more pronounced, which is reflected in the relatively lower IoU values compared to the fall and winter periods. Cyclones, convective weather, and stronger winds can contribute to rapid changes in the positions and types of fronts, which causes the IoU values to fluctuate more and occasionally drop sharply.

Cold fronts and low-pressure systems are typically less frequent and less dynamic during fall and winter months, contributing to more stable front structures. As a result, the IoU increases, and the fronts' positions become more consistent, leading to fewer changes in front IDs from day to day. Large-scale weather patterns such as persistent high-pressure systems or slow-moving cold fronts contribute to the lower variability and higher IoU values in these months.

To sum up, climate and season change plays a vital role in the trend, while weather affects daily IoU values just like brownian motion. This figure proves that fronts are driven by climate, season and weather, and the tracking algorithm is thus validated. 

\begin{figure}
	\centering
	\includegraphics[width=0.35\textwidth]{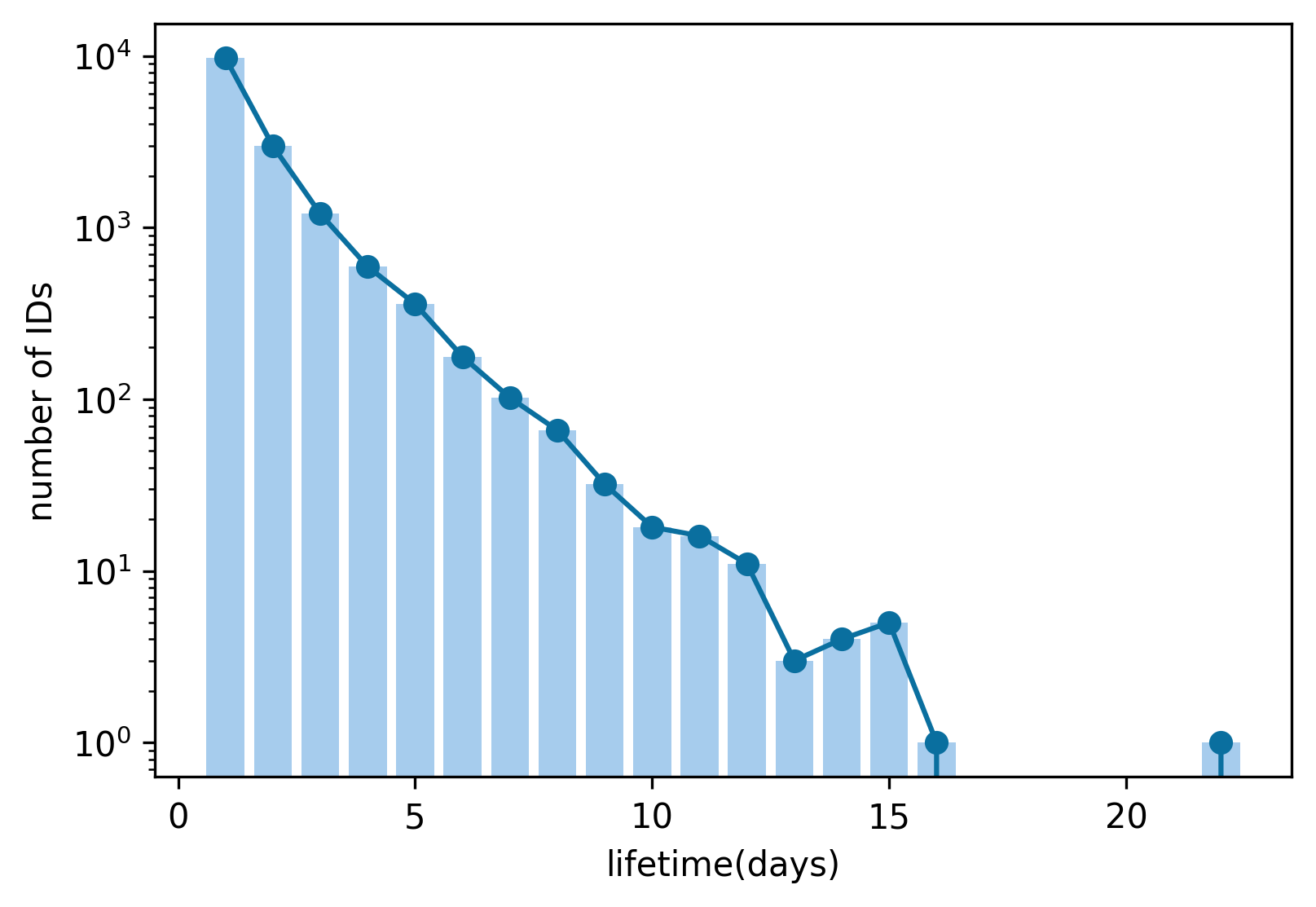}
	\caption{Lifetime distribution of SST front tracking results in 2024} 
	\label{fig:lifetime_stat} 
\end{figure}

The lifetime distribution of SST fronts shows a clear decay trend in the number of fronts as their lifetimes increase, as shown in figure \ref{fig:lifetime_stat}. This aligns with the expectation that many fronts are transient, born and dissipated quickly due to fluctuating environmental conditions. As the lifetime increases beyond 5 days, the number of fronts decreases rapidly, with a few fronts existing for longer durations (up to 15-22 days). This observation is typical for many oceanographic and meteorological phenomena, where short-lived, transient features are much more common than persistent ones.

\begin{figure}
	\centering
	\includegraphics[width=0.35\textwidth]{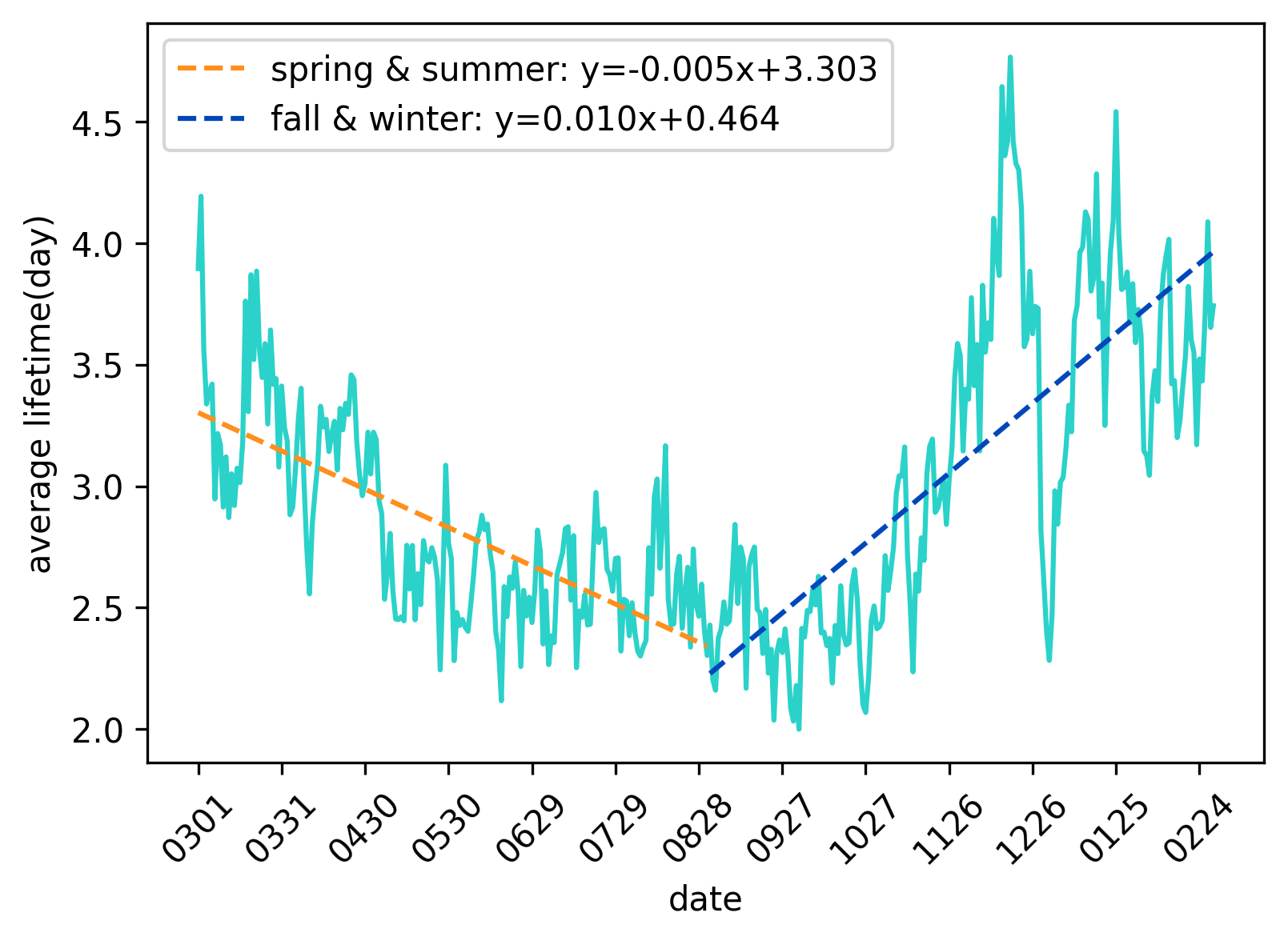}
	\caption{Lifetime of SST front tracking results in 2024} 
	\label{fig:lifetime_reg} 
\end{figure}

Figure \ref{fig:lifetime_reg} provides insight into the temporal variation in the daily average lifetime of SST fronts throughout the year. The decline in average lifetime during spring and summer can be attributed to increased atmospheric instability, as higher temperatures and energy inputs lead to more frequent weather disturbances, resulting in the formation and dissipation of fronts at a faster rate. Conversely, during fall and winter, the lower temperatures and more stable atmospheric conditions contribute to the persistence of SST fronts, leading to longer-lived systems. This also suggests that the frequency of front formation and dissipation is higher in warmer months, which aligns with observations of increased cyclonic activity and fronts' rapid movements.

In comparison to IoU-based analysis (as seen in figure \ref{fig:iou}), which examines daily changes in the spatial consistency of the fronts, the lifetime analysis focuses on the temporal duration of the fronts' existence. The IoU is more sensitive to rapid changes in front positions, such as birth and death events occurring over a short time scale, whereas the lifetime distribution and average lifetime trends are more focused on the overall stability of the fronts over a longer period. This explains why lifetime-based analysis shows lower frequency and less fluctuation compared to IoU-based analysis. While IoU can fluctuate daily, reflecting the instantaneous stability or change in front structure, the lifetime reflects aggregate behavior over time, where fronts that last longer are fewer, leading to a smoother trend in their temporal distribution.

\subsection{Comparison}
To unify the research scale of different methods to $100$km, the window size in histogram method is set at $20$ and the F-test threshold is set at $95\%$. As to LE method, the top $15\%$ FSLE absolute value is selected and MDM is imposed on it to vectorize fronts and generate width data\cite{Sudre-2023}. For the traditional gradient method, the upper and lower threshold is also $10\%$ and $20\%$ as described in \ref{ch:case}. All of the fronts shorter than $100$km is removed. 

\begin{figure*}
	\centering
	\includegraphics[width=0.8\textwidth]{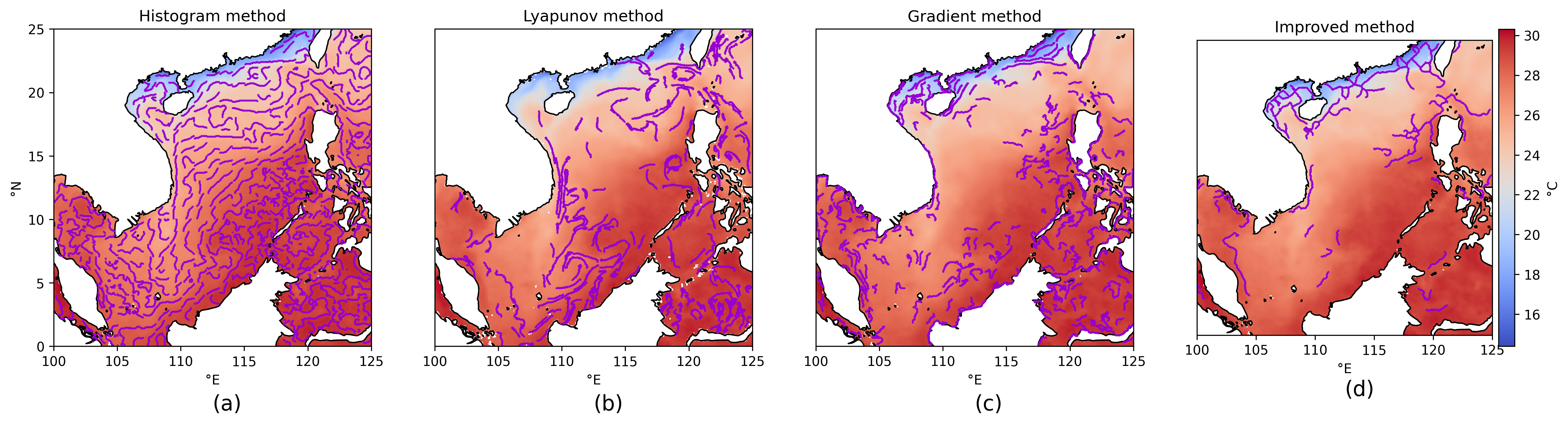}
	\caption{Comparison of SST front detection methods in 2025.01.01} 
	\label{fig:comparison_case} 
\end{figure*}

Figure \ref{fig:comparison_case} compares the results of SST front detection using four different methods. The results are displayed as spatial maps with SST overlaid. It is clear that histogram method displays a large number of detected fronts, but their spatial coherence is poor, leading to potential over-detection. Lyapunov method shows a better representation of front locations compared to the Histogram method, but it still struggles with accurately defining the continuity of the fronts and not fully capturing the front dynamics.Gradient method is better at defining the spatial characteristics of the fronts. However, it still fails to capture the full extent of the fronts and has limitations in detecting less pronounced features.

\begin{table}[]
	\caption{QUANTITATIVE COMPARISONS ON SST DATASET}
	\label{tab:comparison}
	\centering
	\begin{tabular}{lllll}
		\hline
		Feature/Method & Histogram   & Lyapunov   & Gradient & Improved   \\ \hline
		Number       & 263.22 & 228.60  & 421.77   &71.44 \\ 
		Intensity(℃/km)    & 0.091   & 0.066   & 0.093  & 0.16\\
		Width(km)        & --  & 23.75  & --  & 40.56\\
		Length(km)       & 147.53 & 184.02 & 148.37 & 212.42\\ \hline
	\end{tabular}
\end{table}

Table \ref{tab:comparison} presents a quantitative comparison of the four methods, the value is the average of detection results from 2022 to 2024. As to number, improved method significantly outperforms the other methods, detecting only $71.44$ fronts, which indicates a more accurate and less over-detecting performance. In contrast, histogram method detects $263.22$ fronts, indicating that it tends to over-detect front, leading to potential false results. Lyapunov and gradient method also detect more fronts than improved method, but still show better accuracy compared to histogram method. 

Improved method yields the highest intensity of $0.16\degree$C/km, indicating a sharper detection of the fronts. Gradient method comes close with $0.093\degree$C/km, but both histogram and Lyapunov method have much lower intensity values, suggesting weaker and less defined front detection. This suggests that improved method is more capable of detecting stronger fronts, while histogram and Lyapunov methods tend to miss or under-represent the intensity of these fronts.

When it comes to width, improved method achieves the best performance of $40.56$km, demonstrating that it captures more realistic and properly scaled fronts. Lyapunov method ($23.75$km) has much shorter widths, reflecting the influence area is smaller.

Improved method has the highest front length ($212.42$km), indicating that it is better at identifying the full extent of the fronts. In comparison, the other methods have considerably lower length, reflecting that it misses or inaccurately capture the front's full length.

The fundamental reason of over-detection in histogram method lies in the window-based process. Once a window is passed, it generates a dividing SST, ervery point in the window whose SST is larger than dividing SST and surrounding points is marked as a front point. This process just utilizes SST information and the single-threshold feature generates too much front points.

Lyapunov method is based on speed field, and it only detects streamlines around eddies, which is just one of the reasons for front formation. Interaction of ocean and boundary currents, wind stress and Ekman transport, topographic effect, effects of seasonality and climate modes can also generate fronts.

Traditional gradient method is close to the definition of fronts, but only gradient information is used. Lack of merging and ring deletion algorithm results in discontinuity and inaccurate location.

Improved method demonstrates a clear advantage over the other methods in detecting SST fronts. It provides better definitions of front characteristics, and produces more accurate and spatially coherent results. LDE and BD information is added to enhance performance, DSE algorithm is firstly used in oceanography to eliminate trivial branches, merging and ring deletion method are put forward and automatic front tracking algorithm is made public for the first time.

\section{Conclusion}
This paper proposes an automated ocean front detection and tracking framework based on Bayesian decision theory and metric space analysis, addressing key limitations in existing methods such as discontinuity, over-detection, and lack of open-source implementations. By redefining frontal zones as broader candidate regions and integrating gradient information as a prior probability with original field operators, the Bayesian decision mechanism eliminates threshold sensitivity while enhancing detection robustness. Mathematical morphology and ring-deletion algorithms further refine continuity by removing redundant branches and irrational structures. The introduction of metric space for defining temporal front distances enables systematic tracking, a novel contribution to oceanographic research.  

Quantitative comparisons demonstrate the superiority of BFDT-MSA: it reduces over-detection by $73\%$ compared to histogram-based approaches, achieves sharper intensity ($0.16\degree$C/km), and captures broader ($40.56$ km) and longer ($212.42$km) fronts. These improvements stem from the fusion of multi-scale information and iterative refinement strategies, which traditional gradient or Lagrangian methods lack. Additionally, the open-source release of the tracking algorithm fills a critical gap in reproducible research.  

While the framework excels in large-scale front analysis, future work should extend it to submesoscale processes and integrate multi-source data (e.g., salinity, chlorophyll). The methodology's adaptability to edge detection in computer vision suggests broader interdisciplinary applications. Addressing computational efficiency and validating against in-situ observations will further solidify its utility in climate modeling and marine ecosystem management.  

Key innovations: 1. Bayesian fusion: Combines gradient priors with field operators to replace manual thresholds; 2. Morphological refinement: Merging, ring deletion and branch trimming enhance spatial coherence; 3. Metric space tracking: First formal definition of front distance for temporal functional analysis; 4. Open-source accessibility: Public release promotes transparency and community-driven improvements.  

This work bridges methodological gaps in ocean front studies and provides a scalable framework for both operational monitoring and theoretical research.


\bibliographystyle{ieeetr}
\bibliography{reference}

\begin{thebibliography}{10}

\bibitem{Woodson-2015}
C.~B. Woodson and S.~Y. Litvin, ``Ocean fronts drive marine fishery production
  and biogeochemical cycling,'' {\em Proceedings of the National Academy of
  Sciences}, vol.~112, no.~6, pp.~1710--1715, 2015.

\bibitem{Champon-2020}
C.~C. Chapman, M.-A. Lea, A.~Meyer, J.-B. Sall{\'e}e, and M.~Hindell,
  ``Defining southern ocean fronts and their influence on biological and
  physical processes in a changing climate,'' {\em Nature Climate Change},
  vol.~10, no.~3, pp.~209--219, 2020.

\bibitem{Yang-2024}
Y.~Yang, Y.~Ju, Y.~Gao, C.~Zhang, and K.-M. Lam, ``Remote sensing insights into
  ocean fronts: a literature review,'' {\em Intelligent Marine Technology and
  Systems}, vol.~2, no.~1, p.~10, 2024.

\bibitem{Cayula-1992}
J.-F. Cayula and P.~Cornillon, ``Edge detection algorithm for sst images,''
  {\em Journal of atmospheric and oceanic technology}, vol.~9, no.~1,
  pp.~67--80, 1992.

\bibitem{Xing-2024}
Q.~Xing, H.~Yu, and H.~Wang, ``Global mapping and evolution of persistent
  fronts in large marine ecosystems over the past 40 years,'' {\em Nature
  Communications}, vol.~15, no.~1, p.~4090, 2024.

\bibitem{Nieto-2012}
K.~Nieto, H.~Demarcq, and S.~McClatchie, ``Mesoscale frontal structures in the
  canary upwelling system: New front and filament detection algorithms applied
  to spatial and temporal patterns,'' {\em Remote Sensing of Environment},
  vol.~123, pp.~339--346, 2012.

\bibitem{Xing-2023}
Q.~Xing, H.~Yu, H.~Wang, and S.-i. Ito, ``An improved algorithm for detecting
  mesoscale ocean fronts from satellite observations: Detailed mapping of
  persistent fronts around the china seas and their long-term trends,'' {\em
  Remote Sensing of Environment}, vol.~294, p.~113627, 2023.

\bibitem{Yang-2016}
Y.~Yang, J.~Dong, X.~Sun, R.~Lguensat, M.~Jian, and X.~Wang, ``Ocean front
  detection from instant remote sensing sst images,'' {\em IEEE Geoscience and
  Remote Sensing Letters}, vol.~13, no.~12, pp.~1960--1964, 2016.

\bibitem{Aurell-1997}
E.~Aurell, G.~Boffetta, A.~Crisanti, G.~Paladin, and A.~Vulpiani,
  ``Predictability in the large: an extension of the concept of lyapunov
  exponent,'' {\em Journal of physics A: Mathematical and general}, vol.~30,
  no.~1, p.~1, 1997.

\bibitem{Prants-2014}
S.~Prants, M.~Budyansky, and M.~Y. Uleysky, ``Lagrangian fronts in the ocean,''
  {\em Izvestiya, Atmospheric and Oceanic Physics}, vol.~50, pp.~284--291,
  2014.

\bibitem{Boffetta-2001}
G.~Boffetta, G.~Lacorata, G.~Redaelli, and A.~Vulpiani, ``Detecting barriers to
  transport: a review of different techniques,'' {\em Physica D: Nonlinear
  Phenomena}, vol.~159, no.~1-2, pp.~58--70, 2001.

\bibitem{Prants-2011}
S.~Prants, M.~Budyansky, V.~Ponomarev, and M.~Y. Uleysky, ``Lagrangian study of
  transport and mixing in a mesoscale eddy street,'' {\em Ocean modelling},
  vol.~38, no.~1-2, pp.~114--125, 2011.

\bibitem{Prants-2022}
S.~Prants, ``Marine life at lagrangian fronts,'' {\em Progress in
  Oceanography}, vol.~204, p.~102790, 2022.

\bibitem{Tamim-2015}
A.~Tamim, H.~Yahia, K.~Daoudi, K.~Minaoui, A.~Atillah, D.~Aboutajdine, and
  M.~F. Smiej, ``Detection of moroccan coastal upwelling fronts in sst images
  using the microcanonical multiscale formalism,'' {\em Pattern Recognition
  Letters}, vol.~55, pp.~28--33, 2015.

\bibitem{Ma-2024}
Y.~Ma, W.~Liu, B.~Huang, F.~Tian, and G.~Chen, ``Efgan: An automatic gan-based
  methodology for mining eddy-front coupling with fused remote sensing data,''
  {\em Information Fusion}, vol.~101, p.~101982, 2024.

\bibitem{Felt-2022}
V.~C. Felt, {\em Machine learning models for on-orbit detection of temperature
  and chlorophyll ocean fronts}.
\newblock PhD thesis, Massachusetts Institute of Technology, 2022.

\bibitem{Felt-2023}
V.~Felt, S.~Kacker, J.~Kusters, J.~Pendergrast, and K.~Cahoy, ``Fast ocean
  front detection using deep learning edge detection models,'' {\em IEEE
  Transactions on Geoscience and Remote Sensing}, vol.~61, pp.~1--12, 2023.

\bibitem{Yang-2022}
Y.~Yang, K.-M. Lam, X.~Sun, J.~Dong, and R.~Lguensat, ``An efficient algorithm
  for ocean-front evolution trend recognition,'' {\em Remote Sensing}, vol.~14,
  no.~2, p.~259, 2022.

\bibitem{Thomas-2021}
S.~D. Thomas, D.~C. Jones, A.~Faul, E.~Mackie, and E.~Pauthenet, ``Defining
  southern ocean fronts using unsupervised classification,'' {\em Ocean
  Science}, vol.~17, no.~6, pp.~1545--1562, 2021.

\bibitem{Canny-1986}
J.~Canny, ``A computational approach to edge detection,'' {\em IEEE
  Transactions on pattern analysis and machine intelligence}, no.~6,
  pp.~679--698, 1986.

\bibitem{Belkin-2009}
I.~M. Belkin and J.~E. O'Reilly, ``An algorithm for oceanic front detection in
  chlorophyll and sst satellite imagery,'' {\em Journal of Marine Systems},
  vol.~78, no.~3, pp.~319--326, 2009.

\bibitem{Ren-2021}
S.~Ren, X.~Zhu, M.~Drevillon, H.~Wang, Y.~Zhang, Z.~Zu, and A.~Li, ``Detection
  of sst fronts from a high-resolution model and its preliminary results in the
  south china sea,'' {\em Journal of Atmospheric and Oceanic Technology},
  vol.~38, no.~2, pp.~387--403, 2021.

\bibitem{Artal-2019}
O.~Artal, H.~H. Sep{\'u}lveda, D.~Mery, and C.~Pieringer, ``Detecting and
  characterizing upwelling filaments in a numerical ocean model,'' {\em
  Computers \& Geosciences}, vol.~122, pp.~25--34, 2019.

\bibitem{Oram-2008}
J.~J. Oram, J.~C. McWilliams, and K.~D. Stolzenbach, ``Gradient-based edge
  detection and feature classification of sea-surface images of the southern
  california bight,'' {\em Remote Sensing of Environment}, vol.~112, no.~5,
  pp.~2397--2415, 2008.

\bibitem{Bauer-1996}
P.~Bauer, U.~Bodenhofer, and E.~P. Klement, ``A fuzzy algorithm for pixel
  classification based on the discrepancy norm,'' in {\em Proceedings of IEEE
  5th International Fuzzy Systems}, vol.~3, pp.~2007--2012, IEEE, 1996.

\bibitem{Mansoori-2006}
E.~G. Mansoori and H.~J. Eghbali, ``Heuristic edge detection using fuzzy
  rule-based classifier,'' {\em Journal of Intelligent \& Fuzzy Systems},
  vol.~17, no.~5, pp.~457--469, 2006.

\bibitem{Ping-2014}
B.~Ping, F.~Su, Y.~Meng, S.~Fang, and Y.~Du, ``A model of sea surface
  temperature front detection based on a threshold interval,'' {\em Acta
  Oceanologica Sinica}, vol.~33, pp.~65--71, 2014.

\bibitem{Bai-2007}
X.~Bai, L.~J. Latecki, and W.-Y. Liu, ``Skeleton pruning by contour
  partitioning with discrete curve evolution,'' {\em IEEE transactions on
  pattern analysis and machine intelligence}, vol.~29, no.~3, pp.~449--462,
  2007.

\bibitem{Vincent-1993}
L.~Vincent, ``Morphological grayscale reconstruction in image analysis:
  applications and efficient algorithms,'' {\em IEEE transactions on image
  processing}, vol.~2, no.~2, pp.~176--201, 1993.

\bibitem{Hernandez-2011}
I.~Hern{\'a}ndez-Carrasco, C.~L{\'o}pez, E.~Hern{\'a}ndez-Garc{\'\i}a, and
  A.~Turiel, ``How reliable are finite-size lyapunov exponents for the
  assessment of ocean dynamics?,'' {\em Ocean Modelling}, vol.~36, no.~3-4,
  pp.~208--218, 2011.

\bibitem{Sudre-2023}
F.~Sudre, I.~Hern{\'a}ndez-Carrasco, C.~Mazoyer, J.~Sudre, B.~Dewitte,
  V.~Gar{\c{c}}on, and V.~Rossi, ``An ocean front dataset for the mediterranean
  sea and southwest indian ocean,'' {\em Scientific Data}, vol.~10, no.~1,
  p.~730, 2023.

\end{thebibliography}

\end{document}